\documentclass[letterpaper]{article} 
\usepackage{aaai25}  
\usepackage{times}  
\usepackage{helvet}  
\usepackage{courier}  
\usepackage[hyphens]{url}  
\usepackage{graphicx} 
\usepackage{natbib}  
\usepackage{caption} 
\usepackage{floatrow}

\usepackage{graphicx}
\usepackage{amsmath}
\usepackage{amssymb}
\usepackage{mathrsfs} 
\usepackage{tikz}
\usepackage{pgfplots}
\usepackage{adjustbox}
\usetikzlibrary{patterns}

\usepgfplotslibrary{groupplots}

\pgfplotsset{compat=1.18}

\usepackage[linesnumbered,ruled,vlined]{algorithm2e}

\usepackage[utf8]{inputenc} 
\usepackage[T1]{fontenc}    
\usepackage{url}            
\usepackage{booktabs}       
\usepackage{amsfonts}       
\usepackage{nicefrac}       
\usepackage{microtype}      

\usepackage{algorithmic}
\usepackage{subcaption}
\usepackage{xcolor}
\usepackage{bm}
\usepackage{mathtools}
\usepackage{adjustbox}
\usepackage{multirow}

\usepackage{color, colortbl}

\usepackage{helvet} 
\usepackage{courier} 
\usepackage{caption}
\DeclareCaptionStyle{ruled}{labelfont=normalfont,labelsep=colon,strut=off} 

\usepackage{amstext}
\usepackage{enumerate}

\definecolor{amethyst}{rgb}{0.6, 0.4, 0.8}
\definecolor{beaublue}{rgb}{0.74, 0.83, 0.9}
\definecolor{babyblueeyes}{rgb}{0.63, 0.79, 0.95}
\definecolor{babyblue}{rgb}{0.54, 0.81, 0.94}

\definecolor{mygray}{gray}{0.9}

\title{Gradient Weight-normalized Low-rank Projection for Efficient LLM Training}
\author{
  Jia-Hong Huang\textsuperscript{\rm $\dagger$},
  Yixian Shen\textsuperscript{\rm $\dagger$},
  Hongyi Zhu,
  Stevan Rudinac,
  Evangelos Kanoulas\\
}
\affiliations{
    University of Amsterdam \\
    \textsuperscript{\rm $\dagger$}Authors contributed equally to this work \\
    j.huang@uva.nl, y.shen@uva.nl, h.zhu@uva.nl, s.rudinac@uva.nl, E.Kanoulas@uva.nl \\

}

\usepackage{bibentry}

\begin{document}

\maketitle

\begin{abstract}
Large Language Models (LLMs) have shown remarkable performance across various tasks, but the escalating demands on computational resources pose significant challenges, particularly in the extensive utilization of full fine-tuning for downstream tasks. To address this, parameter-efficient fine-tuning (PEFT) methods have been developed, but they often underperform compared to full fine-tuning and struggle with memory efficiency. In this work, we introduce Gradient Weight-Normalized Low-Rank Projection (GradNormLoRP), a novel approach that enhances both parameter and memory efficiency while maintaining comparable performance to full fine-tuning. GradNormLoRP normalizes the weight matrix to improve gradient conditioning, facilitating better convergence during optimization. Additionally, it applies low-rank approximations to the weight and gradient matrices, significantly reducing memory usage during training. Extensive experiments demonstrate that our 8-bit GradNormLoRP reduces optimizer memory usage by up to 89.5\% and enables the pre-training of large LLMs, such as LLaMA 7B, on consumer-level GPUs like the NVIDIA RTX 4090, without additional inference costs. Moreover, GradNormLoRP outperforms existing low-rank methods in fine-tuning tasks. For instance, when fine-tuning the RoBERTa model on all GLUE tasks with a rank of 8, GradNormLoRP achieves an average score of 80.65, surpassing LoRA's score of 79.23. These results underscore GradNormLoRP as a promising alternative for efficient LLM pre-training and fine-tuning.  
Source code: https://github.com/Jhhuangkay/Gradient-Weight-normalized-Low-rank-Projection-for-Efficient-LLM-Training

\end{abstract}

\section{Introduction}
\label{introduction:intro}

Large Language Models (LLMs) pre-trained on extensive datasets have demonstrated exceptional effectiveness across various domains \cite{devlin-etal-2019-bert,liu2019roberta,he2022masked,xie2022simmim,baevski2020wav2vec,lu2019vilbert,tan-bansal-2019-lxmert}. As time progresses, open-source LLMs have consistently improved in their capabilities, accompanied by a striking increase in the scale of pre-trained models \cite{raffel2020exploring,zhang2022opt,le2023bloom,touvron2023llama,tay2023ul2}. 
Consequently, employing full fine-tuning, where all learnable parameters of a pre-trained model are updated for performing downstream tasks, poses unparalleled challenges despite its track record of delivering numerous state-of-the-art results. These challenges primarily stem from the escalating demands on computational resources.

\vspace{-0.4cm}
\begin{algorithm}
\scriptsize
\caption{\small Our proposed GradNormLoRP}
\label{alg:gradnormlorp}
\begin{algorithmic}[1]
\REQUIRE Weight matrix $\mathcal{W}$
\ENSURE Updated weight matrix $\mathcal{W}_{\text{updated}}$
\STATE Normalize each column weight vector of $\mathcal{W}$ to get $\mathcal{W}_{\text{norm}}$
\STATE Apply LoRA with two low-rank matrices $I$ and $J$ to reformulate $\mathcal{W}_{\text{norm}}$
\STATE Initialize two sets of low-rank projection matrices $(U_I, V_I)$ and $(U_J, V_J)$
\FOR{$i=1$ to $N$}
    \STATE Compute gradient matrices $\mathcal{Z}_I$ and $\mathcal{H}_J$ based on $I$ and $J$
    \STATE Project $\mathcal{Z}_I$ and $\mathcal{H}_J$ using $(U_I, V_I)$ and $(U_J, V_J)$
    \IF{$i$ is a multiple of 250}
        \STATE Update $(U_I, V_I)$ and $(U_J, V_J)$
    \ENDIF
\ENDFOR
\STATE \textbf{return} $\mathcal{W}_{\text{updated}}(\mathcal{W}_{\text{norm}}$, $\mathcal{Z}_I$, $(U_I, V_I)$, $\mathcal{H}_J$, $(U_J, V_J))$

\end{algorithmic}
\end{algorithm}
\vspace{-0.4cm}

To tackle the aforementioned challenge, researchers have developed parameter-efficient fine-tuning (PEFT) techniques \cite{houlsby2019parameter,hu2022lora,lialin2023relora,liu2024dora,kopiczko2024vera}. These methods are tailored to update only a small amount of task-specific parameters while leaving the majority of the model's parameters unchanged. Among these techniques, low-rank approximation-based approaches utilize low-rank matrices to approximate weight changes during training, achieving both parameter and memory efficiency without requiring additional trainable subnetworks to be added to the original model architecture. 


Despite their advantages, low-rank-based methods often underperform compared to full-rank fine-tuning \cite{hu2022lora, lialin2023relora, liu2024dora}. This performance gap is typically attributed to the reduced number of trainable parameters, but other underlying factors, such as altered gradient dynamics due to reparameterization, also play a significant role. 
In Figure \ref{fig:figure0} of our \textbf{Appendix}$^1$, we observe that the gradient descent process can become neither smooth nor stable when fine-tuning LLMs in an unnormalized subspace. This instability arises from conducting gradient descent on an incomparable scale, where some values are excessively large or small. Such numerical instability can lead to overflow or underflow during computations, negatively impacting the optimization process and resulting in suboptimal performance. 
To mitigate this problem, we propose our method, Gradient Weight-Normalized Low-Rank Projection (GradNormLoRP). This approach effectively enhances both parameter and memory efficiency. GradNormLoRP improves parameter efficiency by incorporating a weight matrix normalization process that represents each column vector of the weight matrix as the product of its magnitude and unit vector. This normalization enhances gradient conditioning and facilitates better convergence during optimization. 

In addition to improving parameter efficiency, GradNormLoRP addresses memory efficiency while maintaining performance comparable to full fine-tuning without introducing additional inference burden. Although training LLMs in a normalized subspace enhances convergence during optimization, existing PEFT methods \cite{hu2022lora,liu2024dora,houlsby2019parameter,pfeiffer2020adapterfusion,li2021prefix,lester2021power,liu2022few} still face limitations in reducing GPU memory usage. Specifically, these methods rely on caching intermediate activations during the forward pass to compute gradients, which remains a significant memory overhead due to the standard backpropagation process.  This inefficiency poses difficulties for training LLMs on a single consumer-level GPU, such as the NVIDIA RTX 4090 with 24GB of memory. 
To address the memory efficiency issue, GradNormLoRP applies a low-rank approximation technique to both the normalized weight matrix and its corresponding gradient matrix. This process involves reformulating the normalized weight matrix as the sum of a fixed pre-trained weight matrix and the product of two low-rank matrices. It also requires computing two sets of low-rank projection matrices to project the gradient matrices derived from these low-rank matrices. These projection matrices are updated periodically, e.g., every 250 iterations, to ensure minimal additional computational overhead over time.
Exploiting this technique, our proposed GradNormLoRP achieves both memory and parameter efficiency during training while further enhancing the convergence process of optimization in the normalized subspace. 

We conduct extensive experiments to demonstrate the effectiveness of our proposed GradNormLoRP in both LLM pre-training and fine-tuning, leveraging the C4 dataset \cite{JMLR:v21:20-074} and the GLUE benchmark \cite{wang2019glue}. GradNormLoRP significantly reduces memory usage in optimizer states by up to 89.5\%, while preserving efficiency and performance during pre-training on the LLaMA 7B \cite{touvron2023llama} architecture with the C4 dataset, comprising up to 10.2 billion tokens. Furthermore, our 8-bit GradNormLoRP achieves additional reductions, cutting optimizer memory by up to 83.7\% and total training memory by 65.2\% compared to a BF16 baseline. 
Remarkably, we demonstrate the feasibility of pre-training the LLaMA 7B model on consumer-level GPUs with 24GB memory, such as the NVIDIA RTX 4090, without necessitating strategies like model parallelism, offloading, or checkpointing. 
In the realm of fine-tuning pre-trained LLMs on GLUE benchmarks, GradNormLoRP proves superior to existing low-rank methods. For example, when fine-tuning the RoBERTaBase model \cite{liu2019roberta} on GLUE tasks with a rank of 8, GradNormLoRP attains an average score of 80.65, outpacing LoRA, which achieves a score of 79.23. 
The effectiveness of GradNormLoRP is also mathematically proved by our proposed \textbf{Theorem 1 \ref{method:thm4}}. 
This highlights GradNormLoRP as a promising alternative to established methodologies within the field.
The main contributions of this paper are as follows:
\begin{itemize}
    \item \textbf{Development of GradNormLoRP for Enhanced LLM Training}: We introduce GradNormLoRP, a novel method designed to improve parameter and memory efficiency during the pre-training and fine-tuning of LLMs. GradNormLoRP enhances gradient conditioning, leading to better convergence during optimization while maintaining performance comparable to full fine-tuning.

    \item \textbf{Memory Efficiency on Consumer-Level GPUs}: GradNormLoRP addresses the memory efficiency limitations of existing PEFT methods. Through the application of low-rank approximation to both normalized weight matrices and their corresponding gradient matrices, the method substantially reduces memory usage in optimizer states, enabling the training of LLMs on consumer-level GPUs without the need for advanced memory management strategies.

    \item \textbf{Empirical and Theoretical Validation of GradNormLoRP's Effectiveness}: The effectiveness of GradNormLoRP is demonstrated through both theoretical analysis and extensive experimental evaluation. We provide a mathematical proof of GradNormLoRP’s effectiveness, further solidifying its potential as a promising alternative to traditional fine-tuning approaches in the LLM domain.

\end{itemize}

\section{Related Work}
\label{related_work}

\noindent\textbf{Parameter-Efficient Fine-Tuning.}
Numerous PEFT methods have emerged to address the computational challenges of fully fine-tuning LLMs. These methods can be grouped into those that increase model complexity and those that maintain or minimally modify the initial architecture. The first group, including methods like \cite{liao2023make,zhao2024dr,houlsby2019parameter,rebuffi2017learning,gomez2017reversible,pfeiffer2020adapterfusion, ruckle2020adapterdrop, li2021prefix,lester2021power,hambardzumyan2021warp,liu2023gpt}, often incorporate trainable adapter layers or optimize input layer activations, which can add inference latency and pose challenges in large-scale, latency-sensitive environments. 
The second group of methods, including \cite{liu2024dora,hu2022lora,lialin2023relora}, utilizes low-rank matrices to approximate weight changes during training. These low-rank matrices are designed to integrate seamlessly with pre-trained weights before inference, ensuring that no additional inference overhead is introduced. 
Our proposed GradNormLoRP belongs to this second category, leveraging the advantages of low-rank approximation methods without introducing extra inference latency.

\vspace{+2pt}\noindent\textbf{Gradient Projection.}
Gradient projection is used for rapid low-rank estimation \cite{chen2015fast,chen2019non,zhao2024galore}. The work in \cite{chen2015fast,chen2019non} treats the objective function as a general non-linear function, analyzing gradients in vector space. GaLore \cite{zhao2024galore}, however, considers the specific structures of gradients in multi-layer neural networks, establishing that gradients tend to become low-rank during training and exhibit specific convergence behaviors. 
Further studies \cite{larsen2022degrees, gurari2018gradient} demonstrate that effective learning often takes place in a low-dimensional subspace, optimizing model weights within this constrained space—a process known as subspace learning.
Our proposed GradNormLoRP advances this concept by operating within a low-dimensional normalized subspace, enhanced by weight matrix normalization. 

\vspace{+2pt}\noindent\textbf{GPU-Memory-Efficient Training.}
Several techniques have been developed to optimize GPU memory utilization during LLM training. Reversible subnetworks \cite{liao2023make, Mangalam_2022_CVPR, zhao2024dr, NIPS2017_f9be311e, kitaev2020reformer} minimize activation memory by recalculating activations during back-propagation. Gradient checkpointing \cite{chen2016training} improves memory efficiency by discarding and later reconstructing some intermediate activations through an additional forward pass. Pruning \cite{frankle2019lottery, pmlr-v119-frankle20a} and knowledge distillation \cite{sanh2020distilbert, hinton2015distilling, pmlr-v97-koratana19a} compress models by removing redundant parameters or transferring distilled knowledge. Using pre-trained models as feature extractors without gradient computation also reduces activation memory \cite{liu2024tuning, sung2022lst}. Quantization reduces optimizer state memory overhead \cite{dettmers20228bit, NEURIPS2023_3122aaa2}. Fused gradient calculation \cite{lv2023parameter} alleviates memory overhead from storing weight gradients, and Adafactor \cite{pmlr-v80-shazeer18a} reduces memory costs by factorizing second-order statistics. 
Unlike these approaches, GradNormLoRP provides optimizers with low-rank gradients directly, eliminating the need for full-rank gradient knowledge.

\section{Methodology}
\label{method}
In this section, we detail the key components of our GradNormLoRP and establish a theorem that theoretically demonstrates the effectiveness of GradNormLoRP in preserving the integrity of training dynamics. Please consult \textbf{Algorithm 1} for a more comprehensive grasp of our GradNormLoRP. 
\subsection{Background}
\noindent\textbf{Weight Vector Normalization.}
\label{preliminary:norm}
Weight vector normalization is a technique that can be employed to expedite the convergence of the stochastic gradient descent optimization process \cite{srebro2005rank,salimans2016weight}.
We consider standard neural networks in which each neuron's computation involves calculating a weighted sum of input features, followed by a component-wise non-linearity:
\begin{equation}
   \scriptsize y = \theta((\sum_{i=1}^{k} w_i a_i) + b) = \theta(\langle w, a \rangle + b),
\end{equation}
where $w \in \mathbb{R}^{k\times 1}$ represents a weight vector, $a \in \mathbb{R}^{k\times 1}$ signifies an input feature vector, $b \in \mathbb{R}$ indicates a bias term, $\langle \cdot, \cdot \rangle$ denotes the inner product, $\theta(\cdot)$ is an component-wise non-linearity, e.g., the logistic activation $ \frac{\exp(\cdot)}{1+\exp(\cdot)}$, and $y$ indicates the scalar output of the neuron. 

After a loss function is associated with one or more neuron outputs, the parameters $w$ and $b$ for each neuron are typically optimized using stochastic gradient descent during the training of such a neural network.
To enhance the convergence of the optimization process, a reparameterization operation is introduced to express each weight vector $w$ in terms of a parameter vector $v$ and a scalar parameter $\delta$: 
\begin{equation}
   \scriptsize w = \delta \frac{v}{\|v\|},
\end{equation}
where $\delta \in \mathbb{R}$ denotes a scalar, $v \in \mathbb{R}^{k\times 1}$, and $\| \cdot \|$ indicates the Euclidean norm. \\
This reparameterization, which decouples the weight vector's norm ($\delta$) from the direction of the weight vector ($\frac{v}{\|v\|}$), fixes the Euclidean norm of the weight vector, yielding $\|w\| = \delta$, which remains independent of the parameter vector $v$.
After employing the reparameterization weight normalization process, we obtain:
\begin{equation}
    \scriptsize y = \theta(\langle w, a \rangle + b) = \theta(\langle \delta \frac{v}{\|v\|}, a \rangle + b).
\end{equation}
Subsequently, the optimization process of stochastic gradient descent is conducted to the new parameters $v$ and $\delta$ instead.
In our proposed GradNormLoRP approach, we conduct the operation of reparameterization weight normalization on each column weight vector of a given weight matrix, resulting in a normalized weight matrix.

\vspace{+2pt}\noindent\textbf{Challenges in Memory Efficiency for PEFT.}
\label{preliminary:memory_PEFT}
As discussed in \cite{raffel2020exploring,zhao2024galore,liao2023make,touvron2023llama}, the primary memory consumption during neural network training is attributed to activations, trainable parameters, and gradients of these parameters, along with optimizer states such as gradient momentum and variance in Adam \cite{kingma2017adam}. In this subsection, we employ a T-layer multilayer perceptron to illustrate the main origin of the memory efficiency issue inherent in low-rank approximation-based PEFT methods.
Consider a T-layer multilayer perceptron: \( h_T = \xi_T (\xi_{T-1}(\ldots(\xi_2(\xi_1(h_0)))\ldots)) \) with \( h_0 \) as the initial input, where the \( t^{\textrm{th}} \) layer \( h_t = \xi_t(h_{t-1}) = \phi_t(W_t h_{t-1}) \) comprises a nonlinear function \( \phi_t \) and a weight matrix \( W_t \), neglecting the bias term for simplicity. Let \(\psi_t = W_t h_{t-1}\). During the process of backpropagation with a loss \(\mathcal{L}\), the gradient of \(W_t\) is computed using the chain rule as:
\begin{equation}
   \scriptsize \frac{\partial \mathcal{L}}{\partial W_t} = \frac{\partial \mathcal{L}}{\partial h_T} \left( \prod_{i=t+1}^{T} \frac{\partial h_i}{\partial \psi_i} \frac{\partial \psi_i}{\partial h_{i-1}} \right) \frac{\partial h_t}{\partial \psi_t} \frac{\partial \psi_t}{\partial W_t} = \frac{\partial \mathcal{L}}{\partial h_T} \left( \prod_{i=t+1}^{T} \phi_i'W_i \right) \phi'_t h_{t-1},
\end{equation}
where  $\frac{\partial h_i}{\partial \psi_i} = \phi'_i$, $\frac{\partial \psi_i}{\partial h_{i-1}} = W_i$, $\frac{\partial h_t}{\partial \psi_t} = \phi'_t$, and $\frac{\partial \psi_t}{\partial W_t} = h_{t-1}$. \\
Since $\phi_t'$ represents the derivative of $\phi_t$ and the computation of $\phi'_t$ relies on $\psi_t$, caching the sequence of activations $\{ \psi_i \}_{i=t}^{T}$ during the forward pass is essential to compute the gradient of $W_t$, even though $\{ W_i \}_{i>t}$ remain frozen.
In contrast to full fine-tuning, existing low-rank approximation-based PEFT methods adjust only a limited number of parameters, resulting in a negligible size of the optimizer state \cite{liu2024dora,hu2022lora,lialin2023relora,kopiczko2024vera}.
Nevertheless, there is no significant reduction in the memory consumption required for activations. Take BERT$_\text{base}$ fine-tuned on the RTE benchmark with a batch size of 64 and sequence length of 512: the PEFT methods still require over 75\% of the activation memory used in full fine-tuning, even though their trainable parameters are reduced to less than 1\% \cite{devlin2018bert,liao2023make,bentivogli2009fifth}.


\subsection{Our Proposed GradNormLoRP}
\noindent\textbf{Gradient Projection.}
\label{preliminary:gradproj}
The efficacy of existing low-rank approximation-based PEFT approaches, such as LoRA \cite{hu2022lora}, often falls short in comparison to full fine-tuning, primarily due to their limited number of trainable parameters and the potential change of gradient training dynamics resulting from the low-rank reparameterization process \cite{xia2024chain,zhao2024galore,kopiczko2024vera}. A promising avenue to mitigate this challenge is through gradient projection techniques \cite{chen2015fast,chen2019non,zhao2024galore}.
The core concept behind gradient projection is to leverage the gradual evolution of the low-rank structure within the gradient of a weight matrix, instead of directly approximating the weight matrix as done in the LoRA method. This principle is grounded on the claim that the gradient tends to exhibit low-rank characteristics as training progresses. In this subsection, we substantiate this claim through rigorous proof. 

\vspace{+2pt}\noindent\textbf{Weight Matrix Updates in Conventional Full-Rank Training.} 
Given $\mathcal{D}_t = -\nabla_\mathcal{W} \mathcal{L}(\mathcal{W}_t) \in \mathbb{R}^{k \times m}$ as the representation of the backpropagated negative gradient matrix at time step $t$, the traditional pre-training weight update with a learning rate $\alpha$ can be expressed as follows:
\begin{equation}
   \scriptsize \mathcal{W}_T = \mathcal{W}_0 + \alpha \sum_{t=0}^{T-1} \widetilde{\mathcal{D}}_t = \mathcal{W}_0 + \alpha \sum_{t=0}^{T-1} \eta_t(\mathcal{D}_t),
\label{eq:full_rank}
\end{equation}
where \( \widetilde{\mathcal{D}}_t \) represents the final processed gradient added to the weight matrix, and \( \eta_t \) denotes a component-wise stateful gradient regularizer, such as Adam. 

\vspace{+2pt}\noindent\textbf{Weight Matrix Updates in Low-Rank Approximation-Based Methods.} 
For a linear layer with a weight matrix $\mathcal{W} \in \mathbb{R}^{k \times m}$, approaches, such as LoRA, which are based on low-rank approximation, leverage the low-rank structure of the update matrix by introducing a low-rank adaptor $IJ$.
\begin{equation}
   \scriptsize \mathcal{W}_T = \mathcal{W}_0 + I_T J_T,
\label{eq:lora}
\end{equation}
where $I \in \mathbb{R}^{k \times r}$, $J \in \mathbb{R}^{r \times m}$, and $r \ll \min(k, m)$. $I$ and $J$ denote the trainable low-rank adaptors, while $\mathcal{W}_0$ stands as a fixed weight matrix, such as a pre-trained weight matrix. 

While low-rank updates are suggested to alleviate memory consumption, there is ongoing debate regarding whether the weight matrix should inherently adopt a low-rank parameterization. This assumption may not hold in various scenarios, such as linear regression. However, the gradient often exhibits low-rank characteristics during training, especially with specific gradient forms and associated network architectures \cite{zhao2024galore}. The proof for \textbf{Lemma 1} is available in our \textbf{Appendix~\ref{appendix:lemma2.4}}. 

\vspace{+2pt}\noindent\textbf{Lemma 1 (Gradient Becoming Low-rank during Training).} 
\label{lemma1:2-1}
\textit{Given $\mathcal{W}_t \in \mathbb{R}^{k \times m}$, where we assume $k \leq m$ without loss of generality. Consider the gradient matrix $\mathcal{D}_t = \mathcal{A} - \mathcal{B}\mathcal{W}_t\mathcal{C}$, where $\mathcal{A}$ denotes a constant matrix, $\mathcal{B}$ and $\mathcal{C}$ both are positive semidefinite (PSD) matrices, and $\mathcal{W}_0$ is randomly initialized.
Then, the gradient in the update of weight matrix $\mathcal{W}_t = \mathcal{W}_{t-1} + \alpha \mathcal{D}_{t-1}$ results in low-rank gradient with high probability:
\begin{equation}
   \scriptsize \textup{stable-rank}(\mathcal{D}_t) \leq 1 + \sum_{i=2}^k O\left( \left(\frac{1 - \alpha \lambda_i \nu_1}{1 - \alpha \lambda_1 \nu_1}\right)^{2t} \right),
\end{equation}
where $\nu_1 = \lambda_{\min}(\mathcal{C})$ is the smallest eigenvalue of $\mathcal{C}$ and $\lambda_1 \leq \ldots \leq \lambda_m$ are eigenvalues of $\mathcal{B}$. 
Moreover, if $\mathcal{C}$ is positive definite, i.e., $\nu_1 > 0$, and $\lambda_2 > \lambda_1$, $\mathcal{D}_t$ converges exponentially to rank-1.}


\vspace{+2pt}\noindent\textbf{Normalization of Weight Matrix.}
\label{method:norm_matrix}
The initial phase of our proposed GradNormLoRP involves normalizing a provided weight matrix $\mathcal{W}$. This normalization entails reparameterizing each column vector of the weight matrix using the operation introduced in section \textit{``Weight Vector Normalization''}. The normalization process of the weight matrix $\mathcal{W} \in \mathbb{R}^{k \times m}$ can be expressed as follows:
\begin{equation}
    \scriptsize \mathcal{W} = ||\mathcal{W}||_c \frac{\mathcal{W}}{||\mathcal{W}||_c} = \mathcal{M} \frac{\mathcal{W}}{||\mathcal{W}||_c},
\label{eq:norm}
\end{equation}
where $\mathcal{M} \in \mathbb{R}^{1 \times m}$ indicates the reparameterized, i.e., trainable, length vector, $\mathcal{W} / ||\mathcal{W}||_c \in \mathbb{R}^{k \times m}$ represents the directional matrix, and $|| \cdot ||_c$ denotes the vector-wise matrix norm operated across each column. \\
After performing the reparameterization weight normalization operation column-wise on the weight matrix, we have disentangled the magnitude of the weight vectors from their direction. This process ensures that each column of $\mathcal{W} / ||\mathcal{W}||_c$ becomes a unit vector with an associated scalar. Each scalar element in vector $\mathcal{M}$ represents the length of a corresponding vector in weight matrix $\mathcal{W}$.

\vspace{+2pt}\noindent\textbf{Low-rank Approximation.}
\label{method:lora}
The proposed GradNormLoRP is initialized with pre-trained weight $\mathcal{W}_0$ as shown in Equation (\ref{eq:norm}), where $\mathcal{M} = ||\mathcal{W}_0||_c$ and $\mathcal{W} = \mathcal{W}_0$ after initialization. Subsequently, we freeze $\mathcal{W}$ while making $\mathcal{M}$ serve as a trainable vector. The directional matrix is then updated using low-rank approximation techniques, such as LoRA. GradNormLoRP can be formulated similarly to Equation (\ref{eq:lora}) as follows:
\begin{equation}
    \scriptsize \mathcal{W} = \mathcal{M} \frac{\mathcal{W}_0 + IJ}{||\mathcal{W}_0 + IJ||_c},
\label{eq:normij}
\end{equation}
where $\mathcal{M}$ represents a vector comprising trainable parameters, while the weight matrices $I \in \mathbb{R}^{k \times r}$ and $J \in \mathbb{R}^{r \times m}$ are initialized following LoRA's approach to guarantee that $\mathcal{W}$ equals $\mathcal{W}_0$ before fine-tuning. \\ 
As the introduced low-rank approximation in our GradNormLoRP can be merged with the pre-trained weight before inference, it does not introduce any additional latency in the inference phase.

\vspace{+2pt}\noindent\textbf{Gradient Projection  Process.}
\label{method:gradproj}
To enhance the convergence of the optimization process while simultaneously reducing memory usage during training, we integrate the gradient projection technique introduced in section \textit{``Gradient Projection''} into our proposed GradNormLoRP. 

\vspace{+2pt}\noindent\textbf{Singular Value Decomposition (SVD) and Projection Matrices.}
In this study, we utilize SVD to obtain projection matrices that serve the purpose of gradient projection for the gradient matrix $\mathcal{D}_t$:
\begin{equation}
    \scriptsize \mathcal{D}_t = USV^\top \approx \sum_{i=1}^r s_i u_i v_i^\top. 
\label{eq:svd}
\end{equation}
Let $\scriptsize \mathcal{U}_t = [u_1, u_2, \dots, u_r]$ and $\scriptsize \mathcal{V}_t = [v_1, v_2, \dots, v_r]$ denote projection matrices. Then,  $\tilde{\mathcal{D}}_t$ in Equation (\ref{eq:full_rank}) can be expressed as follows: 
\begin{equation}
\scriptsize \tilde{\mathcal{D}}_t = \mathcal{U}_t \eta_t(\mathcal{U}_t^\top \mathcal{D}_t \mathcal{V}_t) \mathcal{V}_t^\top.
\end{equation}
As per \textbf{Lemma 1 \ref{lemma1:2-1}}, the gradient $\mathcal{D}$ may exhibit a low-rank structure. Therefore, by preserving the gradient statistics of a compact ``key portion'' of $\mathcal{D}$ in optimizer states instead of $\mathcal{D}$ itself, significant reductions in memory consumption can be achieved. This motivates the gradient projection strategy integrated into our proposed GradNormLoRP. 

\vspace{+2pt}\noindent\textbf{Definition 1 (Gradient Projection in GradNormLoRP).} 
\textit{The gradient projection strategy in our proposed GradNormLoRP, with a learning rate $\alpha$, follows these gradient update rules:
\begin{equation}
    \scriptsize \mathcal{W}_t = \mathcal{W}_0 + \alpha \sum_{t=0}^{T-1} \tilde{\mathcal{Z}}_t \tilde{\mathcal{H}}_t, \tilde{\mathcal{Z}}_t = P_t \eta_t(P_t^\top \mathcal{Z}_t Q_t) Q_t^\top, \tilde{\mathcal{H}}_t = \mathcal{P}_t \eta_t(\mathcal{P}_t^\top \mathcal{H}_t \mathcal{Q}_t) \mathcal{Q}_t^\top,
\end{equation}
where $\mathcal{W}_0 \in \mathbb{R}^{k \times m}$ denotes the initial weight matrix; $\mathcal{Z}_t \in \mathbb{R}^{k \times r}$ and $\mathcal{H}_t \in \mathbb{R}^{r \times m}$ are the low-rank gradient matrices of the weight matrices $I_{t}$ and $J_{t}$ in Equation (\ref{eq:normij}), respectively. $P_t \in \mathbb{R}^{k \times r}$, $Q_t \in \mathbb{R}^{r \times m}$, $\mathcal{P}_t \in \mathbb{R}^{r \times h}$, and $ \mathcal{Q}_t \in \mathbb{R}^{m \times s}$ are projection matrices.} \\
In contrast to LoRA, our proposed GradNormLoRP adopts a distinct approach by employing two low-rank updates, i.e., $\tilde{\mathcal{Z}}_t$ and $\tilde{\mathcal{H}}_t$, explicitly, avoiding the introduction of additional low-rank adaptors and thereby mitigating the alteration of training dynamics. 
Essentially, integrating GradNormLoRP into model training facilitates smooth transitions across normalized low-rank subspaces, as delineated in Equation (\ref{eq:subspaces}).
This means the model can smoothly navigate between different sets of parameters, akin to switching lanes on a highway to optimize its learning process.
\begin{equation}
    \scriptsize \mathcal{W}_t = \mathcal{W}_0 + \Delta \mathcal{W}_{T_1} + \Delta \mathcal{W}_{T_2} + \ldots + \Delta \mathcal{W}_{T_m},
\label{eq:subspaces}
\end{equation}
where $t \in [\sum_{i=1}^{m-1} T_i,\sum_{i=1}^{m} T_i]$ and $\Delta \mathcal{W}_{T_i} = \alpha \sum_{t=0}^{T_{i}-1} \tilde{\mathcal{D}}_t$ denotes the sum of all $T_i$ updates within the $i$-th normalized subspace. 

The effectiveness of GradNormLoRP hinges on the premise that gradients often exhibit low-rank properties throughout training. To validate this assertion, we present \textbf{Theorem 1 \ref{method:thm4}}, with its proof provided in our \textbf{Appendix}$^1$.

\vspace{+2pt}\noindent\textbf{Theorem 1}
\label{method:thm4}
\textit{Let \( r \leq m \) without loss of generality. 
The gradient update rules of GradNormLoRP:
\begin{equation}
   \scriptsize \mathcal{Z}_t = A - BI_tC,~  I_t = I_{t-1} + \gamma \mathcal{Z}_{t-1}~~;~~
\mathcal{H}_t = E - FJ_tG,~  J_t = J_{t-1} + \beta \mathcal{H}_{t-1},
\end{equation}
with constant matrices ($A$ and $E$), PSD matrices ($B$, $C$, $F$, and $G$), and randomly initialized $I_0$ and $J_0$ leads to low-rank gradient with high probability:
\begin{equation}
\scriptsize \textup{stable-rank}(\mathcal{Z}_t, \mathcal{H}_t) \leq 1 +  \sum_{i=2}^r O\left( \left(\frac{1 - \gamma \omega_i \nu_1}{1 - \gamma \omega_1 \nu_1}\right)^{2t} \right) \sum_{j=2}^r O\left( \left(\frac{1 - \beta \pi_j \mu_1}{1 - \beta \pi_1 \mu_1}\right)^{2t} \right),
\end{equation}}

\section{Experiments}

In this section, we evaluate the efficacy of our proposed GradNormLoRP through a series of experiments. We assess its performance in fine-tuning and pre-training scenarios and conduct a thorough throughput analysis to confirm that GradNormLoRP integrates seamlessly without adding inference latency. Additionally, we perform comprehensive ablation studies to highlight GradNormLoRP's characteristics, including convergence speed, parameter efficiency, and GPU memory utilization.
\subsection{Experimental Setup}

\noindent\textbf{Datasets, Evaluation Metrics, Model Architectures, and Baselines.} 
For fine-tuning, we use the GLUE benchmark \cite{wang2019glue}, which includes single-sentence tasks (CoLA, SST-2), similarity and paraphrase tasks (MRPC, QQP, STS-B), and inference tasks (MNLI, QNLI, RTE, WNLI). Evaluation metrics are accuracy for MNLI, QQP, QNLI, SST-2, MRPC, and RTE, Pearson and Spearman correlation for STS-B, and Matthews correlation for CoLA. For pre-training, we use the C4 dataset \cite{JMLR:v21:20-074}, a cleaned version of Common Crawl's web corpus, with perplexity as the performance metric. \\
In our fine-tuning experiments, we use BERT${\textup{base}}$ \cite{devlin2018bert}, RoBERTa${\textup{base}}$, RoBERTa${\textup{large}}$ \cite{liu2019roberta}, and BART${\textup{base}}$ \cite{lewis2019bart} for all GLUE tasks. For pre-training, we adopt the LLaMA model architecture, training on the C4 dataset with no data repetition, scaling up to 7 billion parameters. \\
Our primary baseline for fine-tuning is full parameter updating. For PEFT experiments, we compare GradNormLoRP against LoRA \cite{hu2022lora}, DoRA \cite{liu2024dora}, and GaLore \cite{zhao2024galore}, which are also used as baselines for our pre-training experiments due to their relevance in low-rank approximation methods.

\vspace{+2pt}\noindent\textbf{Implementation.}
For fine-tuning, we evaluated our model on the GLUE benchmark, exploring learning rates in the range of \{1e-4, 2e-4, 3e-4, 4e-4, 5e-4\}, batch sizes of 16 and 32, and a fixed number of 30 epochs. Specifically, we used a batch size of 16 for all tasks except for CoLA, which used a batch size of 32. The maximum sequence length for all tasks was set to 512 for BERT$_{\textup{base}}$, RoBERTa$_{\textup{base}}$, RoBERTa$_{\textup{large}}$, and BART$_{\textup{base}}$ models. For pretraining, we applied GradNormLoRP across various model sizes ranging from 60M to 1B parameters. The hyperparameters for GradNormLoRP were consistent across all models, with a learning rate of 0.01 and a scale factor ($\alpha_s$) of 0.25. The learning rate was fine-tuned from the set \{1e-2, 1e-3,5e-4,1e-4\}, selecting the best rate based on validation perplexity. Each model was pre-trained for 10,000 steps. For models scaled up to 7B parameters, we set the batch size to 16 and varied the training steps accordingly.



\begin{table*}[t!]
\small
\centering
\scalebox{0.75}{
\begin{tabular}{c|cc|ccccccccccc}
\toprule
\multicolumn{2}{c|}{\textbf{Model}}  & \textbf{Memory} &\textbf{CoLA} & \textbf{MNLI} & \textbf{MRPC} & \textbf{QNLI} & \textbf{QQP} & \textbf{RTE} & \textbf{SST-2} & \textbf{STS-B} & \textbf{WNLI} & \textbf{Avg}\\ 
\midrule
\multicolumn{2}{c|}{Full FT} &747M  &57.03 & 86.30 & 92.09 & 92.04 & 91.35 & 77.62 & 91.74 & 90.82 & 43.90  & 80.32  \\ 
\midrule
\multicolumn{2}{c|}{LoRA (r=4)} & 257M     &55.27	& 86.81 & 89.95	& 92.42	& 89.44	& 69.68 & 93.58	& 90.07 & \textbf{56.34}  & 80.40 \\ 
\multicolumn{2}{c|}{DoRA (r=4)}  & 259M    & 55.51	& \textbf{86.90} & 89.91	& 92.24	& 89.54	& 70.75 & 93.46	& 90.04 & \textbf{56.34}  & 80.52 \\ 
\multicolumn{2}{c|}{GaLore (r=4)}  & 253M    & \textbf{60.65}	& 85.65 & \textbf{91.14}	& 90.76	& \textbf{90.70}	& \textbf{77.26} & 92.89	& 90.84 & 36.62  & 79.61 \\ 
\rowcolor{gray!20} \multicolumn{2}{c|}{\textbf{Ours (r=4)}} & 249M & 59.31 & 86.42	& 91.10	& \textbf{92.48}	& 90.60 & 75.81	& \textbf{94.50} & \textbf{90.85} & 47.89 & \textbf{81.01}\\ 
\midrule
\multicolumn{2}{c|}{LoRA (r=8)}  &264M    & 56.50	& 86.65 & 90.30	& 92.60	& 89.74	& 69.68 & 93.81	& 90.10 & \textbf{43.67}  & 79.23 \\ 
\multicolumn{2}{c|}{DoRA (r=8)}   & 267M  & 57.27	& 86.55 & 89.56	& 92.46	& 89.86	& 70.04 & 94.15	& 90.16 & 43.66  & 79.30 \\ 
\multicolumn{2}{c|}{GaLore (r=8)}  & 257M & 52.59 & 85.66 & \textbf{92.06} & 91.31 & 90.74 & \textbf{78.70}	& 92.66 & 90.80  & 39.44 & 79.33 \\ 
\rowcolor{gray!20} \multicolumn{2}{c|}{\textbf{Ours (r=8)}} & 251M & \textbf{60.31} & \textbf{86.97}	& 91.36	& \textbf{92.62}	& \textbf{91.01} & 77.26	& \textbf{94.50} & \textbf{90.83}  & 40.85 & \textbf{80.65}\\ 
\bottomrule

\end{tabular}}
\caption{\small Evaluating GradNormLoRP for memory-efficient fine-tuning on the GLUE benchmark using the pre-trained RoBERTa$_\textup{base}$ model. ``r'' indicates rank, ``Ours'' signifies GradNormLoRP, and ``FT'' denotes fine-tuning.}
\label{table:2-1_ft}
\vspace{-0.2cm}
\end{table*}


\begin{table}[t!]
\small
\centering
\scalebox{0.75}{
\begin{tabular}{c|cccccc}
\toprule
\multicolumn{2}{c|}{\textbf{Method}}  &\textbf{60M} & \textbf{130M} & \textbf{350M} & \textbf{1B} \\ 
\midrule
\multicolumn{2}{c|}{Full-rank}  & 34.51;(0.35G) & 25.91;(0.8G) & 20.24;(2.21G) & 16.86;(8.03G)    \\ 
\midrule
\multicolumn{2}{c|}{LoRA}  & 35.33;(0.35G) & 30.55;(0.81G) & 25.11;(1.93G) & 22.35;(6.32G)   \\ 
\multicolumn{2}{c|}{DoRA}  & 35.42;(0.37G) & 30.92;(0.82G) & 24.91;(1.95G) & 21.98;(6.37G) 	\\ 
\multicolumn{2}{c|}{GaLore}  & 34.94;(0.23G) & 26.57;(0.54G) & 20.64;(1.47G) & 16.77;(4.69G)  	\\ 
\rowcolor{gray!20} \multicolumn{2}{c|}{\textbf{GradNormLoRP}}  & \textbf{34.63};(0.17G) & \textbf{26.52};(0.47G) & \textbf{19.28};(1.19G) & \textbf{16.12};(3.81G)  	\\ 
\midrule
\multicolumn{2}{c|}{$r/d_{model}$}  & 128 / 256 & 256 / 768 & 256 / 1024 & 512 / 2048 	\\ 
\multicolumn{2}{c|}{Training Tokens}  & 1.1B & 2.2B & 6.4B & 13.1B 	\\ 
\bottomrule
\end{tabular}}
\caption{\footnotesize Compared with low-rank algorithms on pre-training various sizes of LLaMA models on the C4 dataset, reporting validation perplexity and memory estimates for parameters and optimizer states in BF16 format, with actual memory usage as shown.}
\label{table:2-pr}
\vspace{-0.2cm}
\end{table}

\begin{table}[t!]
\small
\centering
\scalebox{0.75}{
\begin{tabular}{c|cc|ccccc}
\toprule
\multicolumn{2}{c|}{\textbf{Model}}  &\textbf{Memory} & \textbf{20K} & \textbf{40K} & \textbf{60K} & \textbf{80K} \\ 
\midrule
\rowcolor{mygray} \multicolumn{2}{c|}{\textbf{8-bit GradNormLoRP}}  &  \textbf{15.29G} & \textbf{19.33} & \textbf{17.73} & \textbf{16.43} & \textbf{15.41}  \\ 
\multicolumn{2}{c|}{8-bit GaLore}  & 18.36G & 20.19 & 18.15 & 16.96&  16.08   \\ 
\multicolumn{2}{c|}{8-bit Adam}  & 26.47G & 20.65 & 18.31& 17.11 &  16.24 	\\ 
\midrule
\multicolumn{2}{c|}{Training Tokens}  & &2.5B  & 5.2B & 7.7B & 10.5B  	\\ 
\bottomrule
\end{tabular}}
\caption{\footnotesize Pre-training LLaMA 7B on the C4 dataset for 80K steps, with validation perplexity and memory estimates reported.}
\label{table:2-spr}
\vspace{-0.2cm}
\end{table}

\subsection{Results and Analysis}
\textbf{Quantitative Results for Fine-tuning.} 
We fine-tune pre-trained RoBERTa models on GLUE tasks using GradNormLoRP and compare its performance with a full fine-tuning baseline, LoRA, DoRA, and GaLore. We use hyperparameters from \cite{hu2022lora} for LoRA and \cite{liu2024dora} for DoRA, and tune the learning rate and scale factor for GradNormLoRP. As shown in Table~\ref{table:2-1_ft}, GradNormLoRP achieves better performance than LoRA and DoRA on most tasks with a lower memory footprint. For instance, GradNormLoRP achieves an average score of 81.01 with a memory usage of 249M for rank=4, while LoRA and DoRA achieve average scores of 80.40 and 80.52 with memory usages of 257M and 259M, respectively. This demonstrates that GradNormLoRP can serve as a full-stack memory-efficient training strategy for fine-tuning. Specifically, GradNormLoRP shows notable improvements in tasks such as SST-2, where it achieves 94.50 compared to 92.89 for GaLore. Additionally, GradNormLoRP maintains competitive performance in other tasks like QNLI and QQP, demonstrating its robustness.

\vspace{+2pt}\noindent\textbf{Quantitative Results for Pre-training.}
For GradNormLoRP and GaLore, we set subspace frequency $T$ to 250 and scale factor $\alpha_s$ to 0.25 across all model sizes in Table~\ref{table:2-pr}. For each model size, we pick the same rank $r$ for all low-rank methods, applying them to all the linear layers of all multi-head attention layers and feed-forward layers in the models. We keep the Adam optimizer settings consistent with GaLore~\cite{zhao2024galore}. We also estimate the memory usage based on BF16 format, including the memory for weight parameters and optimizer states. We evaluate them on LLaMA 60M, 130M, 350M and 1B architecture with 10K training steps, and we tune the learning rate for each setting and report the best performance. \\
As shown in Table~\ref{table:2-pr}, GradNormLoRP achieves significant reductions in validation perplexity and memory usage across different model sizes compared to other methods. For instance, for the 350M parameter model, GradNormLoRP achieves a perplexity of 19.28 with a memory usage of 1.19G, whereas GaLore achieves a perplexity of 20.64 with a memory usage of 1.47G. This represents a substantial improvement in both memory efficiency and model performance. Similarly, for the 1B parameter model, GradNormLoRP reduces the perplexity to 16.12 and memory usage to 3.81G, outperforming GaLore's perplexity of 16.77 and memory usage of 4.69G. These results demonstrate that GradNormLoRP can significantly enhance the efficiency of pre-training LLMs, making it a robust and scalable solution for training in resource-constrained environments. 

\begin{table*}[t!]
\small
\centering
\scalebox{0.75}{
\begin{tabular}{c|cc|ccccccccccc}
\toprule
\multicolumn{2}{c|}{\textbf{Model}}  &\textbf{Memory} & \textbf{CoLA} & \textbf{MNLI} & \textbf{MRPC} & \textbf{QNLI} & \textbf{QQP} & \textbf{RTE} & \textbf{SST-2} & \textbf{STS-B} & \textbf{WNLI} & \textbf{Avg}\\ 
\midrule
\multicolumn{2}{c|}{Full FT$_\textup{RoBERTa}$} &747M  &57.03 & 86.30 & 92.09 & 92.04 & 91.35 & 77.62 & 91.74 & 90.82 & 43.90  & 80.32  \\ 
\multicolumn{2}{c|}{Full FT$_\textup{BART}$}  & 901M & 53.78 & 86.16 & 91.66 & 91.83 & 91.01 & 77.11 & 91.75 & 89.87 & 39.82  & 79.25  \\ 
\multicolumn{2}{c|}{Full FT$_\textup{BERT}$}  & 708M & 56.47 & 86.29 & 91.72 & 91.97 & 91.04 & 77.12 & 91.79 & 89.99 & 39.85  & 79.60   \\ 
\midrule
\rowcolor{mygray} \multicolumn{2}{c|}{Ours$_\textup{RoBERTa}$} & 251M & \textbf{60.31} & \textbf{86.97}	& 91.36	& \textbf{92.62}	& \textbf{91.01} & 77.26	& \textbf{94.50} & \textbf{90.83}  & 40.85 & \textbf{80.65}\\ 
\multicolumn{2}{c|}{Ours$_\textup{BART}$}  & 361M & 54.13 & 86.29  & 91.00 & 91.85 & 90.07 & 73.31 & 93.23 & 89.15 & 49.30  & 79.81  \\ 
\multicolumn{2}{c|}{Ours$_\textup{BERT}$}  & 236M & 58.57 & 86.95 & 89.75 & 90.94 & 91.42 & 70.76 & 92.20 & 88.69 & 44.99  & 79.26   \\ 
\bottomrule
\end{tabular}}
\caption{\footnotesize Model architecture ablation study under the same model size. ``r'' denotes rank, ``Ours'' indicates GradNormLoRP, and ``FT'' signifies fine-tuning. GradNormLoRP utilizes a rank of 8 here.}
\label{table:2-1ma}
\vspace{-0.2cm}
\end{table*}

\vspace{+2pt}\noindent\textbf{Pre-training on LLaMA 7B.}
Scaling to 7B models is crucial for demonstrating GradNormLoRP's effectiveness in practical LLM pre-training. We evaluate GradNormLoRP on an LLaMA 7B architecture, which has an embedding size of 4096 and 32 layers. The model is trained for 80K steps with 10.5B tokens, using 8-node parallel training on 32 A100 GPUs. Due to computational constraints, we compare 8-bit GradNormLoRP (r = 1024) with 8-bit Adam, performing a single trial without hyperparameter tuning. As shown in Table~\ref{table:2-spr}, 8-bit GradNormLoRP not only has a lower memory footprint but also achieves better performance metrics. Specifically, 8-bit GradNormLoRP requires 15.29GB of memory, significantly less than the 18.36GB and 26.47GB required by 8-bit GaLore and 8-bit Adam, respectively. \\
In terms of validation perplexity, 8-bit GradNormLoRP consistently outperforms other methods across training steps, achieving lower perplexity with higher ranks. At 20K steps, 8-bit GradNormLoRP reaches a perplexity of 19.33, compared to 20.19 for 8-bit GaLore and 20.65 for 8-bit Adam. This trend persists at 40K, 60K, and 80K steps, where 8-bit GradNormLoRP achieves 15.41 perplexity, outperforming 8-bit GaLore (16.08) and 8-bit Adam (16.24). Its superior performance stems from efficient low-rank adaptation during gradient projection, optimizing memory usage and enhancing training efficiency, making it a highly effective solution for pre-training LLMs with improved hardware utilization.

\begin{figure}
\vspace{-0.4cm}
    \centering
    \includegraphics[width=\linewidth]{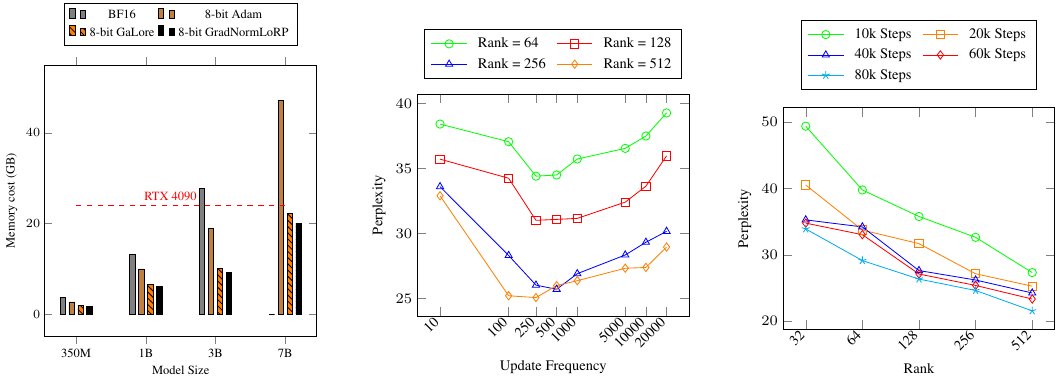}
     \caption{\small From left to right, the figure illustrates a comparison of memory usage, the impact of varying subspace frequencies, and the effect of rank across steps.}
    \label{fig:combined-figures}
    \vspace{-0.3cm}
\end{figure}

Regarding memory consumption, as shown in the left subfigure of Figure~\ref{fig:combined-figures}, 8-bit GradNormLoRP requires only 20.07GB to pre-train the LLaMA 7B model with a per-GPU token batch size of 256, well within the 24GB VRAM of an NVIDIA RTX 4090. This is substantially lower than BF16 and 8-bit Adam, which exceed 24GB for larger models.






\vspace{+2pt}\noindent\textbf{Subspace Update Frequency Ablation Study.} This ablation study, illustrated in the middle subfigure of Figure~\ref{fig:combined-figures}, highlights the importance of finding an optimal update frequency for subspace updates to achieve the best model convergence. Both overly frequent and overly infrequent updates negatively impact performance, leading to increased perplexity. The study shows that the optimal performance occurs at a moderate update frequency of around 250 iterations, especially for higher ranks such as 256 and 512, which benefit from more effective optimization within larger subspaces. Weight normalization plays a crucial role by stabilizing the gradient descent process, ensuring that updates are on a comparable scale and preventing inefficiencies.

\vspace{+2pt}\noindent\textbf{Rank of Subspace Ablation Study.}  
As shown in the right subfigure of Figure~\ref{fig:combined-figures}, this study examines how subspace rank and training steps affect perplexity. Training with rank 128 for 80K steps achieves better perplexity than rank 512 at 20K steps, emphasizing the importance of sufficient training duration for higher ranks. GradNormLoRP excels by effectively combining subspace updates and weight normalization, stabilizing gradient descent. While higher ranks offer significant gains, they require more steps, whereas lower ranks optimize efficiently with fewer steps but show more gradual perplexity improvements.


\vspace{+2pt}\vspace{+2pt}\noindent\textbf{Model Architecture Ablation Study.}  We evaluated GradNormLoRP on BART$\textup{base}$, BERT$\textup{base}$, and RoBERTa$\textup{base}$ to assess its robustness. As shown in Table~\ref{table:2-1ma}, GradNormLoRP consistently outperforms full fine-tuning. For RoBERTa$\textup{base}$ at rank=8, it achieves an average score of 80.65, with significant improvements in CoLA (60.31) and SST-2 (94.50), compared to 80.32 for full fine-tuning. For BART$\textup{base}$, it scores 79.81 on average, excelling in MNLI (86.97) and MRPC (91.06) over full fine-tuning's 79.25. For BERT$\textup{base}$, it maintains a competitive average score of 79.26, with notable results in QNLI (89.75) and QQP (91.42), compared to 79.60 for full fine-tuning. 

\noindent\textbf{Model Size Ablation Study.} We evaluated GradNormLoRP on RoBERTa$\textup{base}$ and RoBERTa$\textup{large}$, demonstrating significant memory savings while maintaining or improving performance compared to full fine-tuning (see Table~\ref{table:2-1} in the \textbf{Appendix}$^1$). For RoBERTa$\textup{base}$ (rank=8), GradNormLoRP achieves an average score of 80.63, with notable gains in CoLA (60.31 vs. 57.03) and SST-2 (94.50 vs. 91.74). For RoBERTa$\textup{large}$, it achieves 82.43, slightly outperforming full fine-tuning (82.39), underscoring GradNormLoRP's efficiency across model sizes.



\vspace{+2pt}\noindent\textbf{Weight Normalization Ablation Study.}  
We conducted an ablation study to evaluate the impact of weight normalization on the RoBERTa$_\textup{base}$ model's performance across GLUE benchmark tasks. As shown in Table~\ref{table:2-1wo} in our \textbf{Appendix}$^1$, weight normalization consistently improves performance. The improvement is most noticeable in tasks like CoLA, where accuracy significantly increases. While gains in SST-2 are minor, they align with the overall positive trend. In more complex tasks like MRPC, weight normalization provides notable benefits, indicating its role in stabilizing and optimizing the training process.


\vspace{+2pt}\noindent\textbf{Gradient Projection Ablation Study.}  
We compared the performance and memory usage of the RoBERTa$_\textup{base}$ model with and without gradient projection across GLUE tasks. As shown in Table~\ref{table:2-1wop} in our \textbf{Appendix}$^1$, incorporating gradient projection in GradNormLoRP significantly boosts performance while preserving memory efficiency. Models with gradient projection demonstrate improved stability and training dynamics, leading to higher scores in tasks like CoLA and SST-2.

\vspace{-0.2cm}
\section{Conclusion}
GradNormLoRP addresses the growing computational demands of full fine-tuning for LLMs by enhancing parameter and memory efficiency while maintaining comparable performance. By normalizing weight matrices, applying low-rank approximation, and utilizing gradient low-rank projection, GradNormLoRP significantly reduces memory overhead during training without adding inference burden. We validate its effectiveness both mathematically and empirically. Our experiments demonstrate its efficacy in LLM pre-training and fine-tuning, achieving comparable or superior results to existing PEFT methods. 

\section{Acknowledgements}
The computational support for this research was provided by the Netherlands Organization for Scientific Research (NWO) under project number EINF-9627.

\clearpage

\bibliography{aaai25}
\clearpage

\newpage
\appendix

\section*{\huge Appendix} 
\label{appendix:1}

\vspace{+0.4cm}
\section{Proof of Our Proposed Theorem 1}

\textbf{Theorem 1 (Low-Rank Evolution of Gradient in GradNormLoRP during Training).}
\textit{Let \( r \leq m \) without loss of generality. 
The gradient update rules of GradNormLoRP:
\begin{equation}
   \tiny \mathcal{Z}_t = A - BI_tC,~  I_t = I_{t-1} + \gamma \mathcal{Z}_{t-1}~~;~~
\mathcal{H}_t = E - FJ_tG,~  J_t = J_{t-1} + \beta \mathcal{H}_{t-1},
\end{equation}
with constant matrices ($A$ and $E$), PSD matrices ($B$, $C$, $F$, and $G$), and randomly initialized $I_0$ and $J_0$ leads to low-rank gradient with high probability:
\begin{equation}
\tiny \textup{stable-rank}(\mathcal{Z}_t, \mathcal{H}_t) \leq 1 +  \sum_{i=2}^r O\left( \left(\frac{1 - \gamma \omega_i \nu_1}{1 - \gamma \omega_1 \nu_1}\right)^{2t} \right) \sum_{j=2}^r O\left( \left(\frac{1 - \beta \pi_j \mu_1}{1 - \beta \pi_1 \mu_1}\right)^{2t} \right)
\end{equation}}
\textbf{\textit{<Proof>}} 
By \textbf{Lemma 1 \ref{lemma1:2-1}}, \( \nu_1 = \omega_{\min}(C) \) denotes the smallest eigenvalue of \( C \), with \( \omega_1 \leq \ldots \leq \omega_r \) representing the eigenvalues of \( B \). Additionally, if \( \omega_2 > \omega_1 \) and \( \nu_1 > 0 \), then \( \mathcal{Z}_t \) converges exponentially to rank-1. Similarly, by \textbf{Lemma 1 \ref{lemma1:2-1}}, \( \mu_1 = \pi_{\min}(G) \) signifies the smallest eigenvalue of \( G \), while \( \pi_1 \leq \ldots \leq \pi_m \) are the eigenvalues of \( F \). Moreover, if \( \pi_2 > \pi_1 \) and \( \mu_1 > 0 \), \( \mathcal{H}_t \) converges exponentially to rank-1. Consequently, the stable rank of $\text{stable-rank}(\mathcal{Z}_t)\text{stable-rank}(\mathcal{H}_t)$, denoted as \( \text{stable-rank}(\mathcal{Z}_t,\mathcal{H}_t) \), converges exponentially to rank-1. $\quad \square$

\section{Proof of Lemma 1}

\noindent\textbf{Lemma 1 (Gradient Becoming Low-rank during Training).} 
\label{appendix:lemma2.4}
\textit{Given $\mathcal{W}_t \in \mathbb{R}^{k \times m}$, where we assume $k \leq m$ without loss of generality. Consider the gradient matrix $\mathcal{D}_t = \mathcal{A} - \mathcal{B}\mathcal{W}_t\mathcal{C}$, where $\mathcal{A}$ denotes a constant matrix, $\mathcal{B}$ and $\mathcal{C}$ both are positive semidefinite (PSD) matrices, and $\mathcal{W}_0$ is randomly initialized.
Then, the gradient in the update of weight matrix $\mathcal{W}_t = \mathcal{W}_{t-1} + \alpha \mathcal{D}_{t-1}$ results in low-rank gradient with high probability:
\begin{equation}
    \textup{stable-rank}(\mathcal{D}_t) \leq 1 + \sum_{i=2}^k O\left( \left(\frac{1 - \alpha \lambda_i \nu_1}{1 - \alpha \lambda_1 \nu_1}\right)^{2t} \right),
\end{equation}
where $\nu_1 = \lambda_{\min}(\mathcal{C})$ is the smallest eigenvalue of $\mathcal{C}$ and $\lambda_1 \leq \ldots \leq \lambda_m$ are eigenvalues of $\mathcal{B}$. 
Moreover, if $\mathcal{C}$ is positive definite, i.e., $\nu_1 > 0$, and $\lambda_2 > \lambda_1$, then $\mathcal{D}_t$ converges exponentially to rank-1.}

\textbf{\textit{<Proof>}} We have
\begin{align}
\mathcal{D}_t &= \mathcal{A} - \mathcal{B} \mathcal{W}_t \mathcal{C} = \mathcal{A} - \mathcal{B} (\mathcal{W}_{t-1} + \eta \mathcal{D}_{t-1}) \mathcal{C} \notag \\
    &= \mathcal{D}_{t-1} - \eta \mathcal{B} \mathcal{D}_{t-1} \mathcal{C}.
\end{align}
Let \( \mathcal{B} = U \tilde{D}_\mathcal{B} U^\top \) and \( \mathcal{C} = V \tilde{D}_\mathcal{C} V^\top \) be the eigen decomposition of \( \mathcal{B} \) and \( \mathcal{C} \). \( \tilde{D}_\mathcal{B} = \text{diag}(\lambda_1, \ldots, \lambda_m) \) and \( \tilde{D}_\mathcal{C} = \text{diag}(\nu_1, \ldots, \nu_n) \) are their eigenvalues sorted in ascending orders (i.e., \( \lambda_1 \leq \ldots \leq \lambda_m \) and \( \nu_1 \leq \ldots \leq \nu_n \)). Define \( \mathcal{X}_t := U^\top \mathcal{D}_t V \). It is clear that \( \text{rank}(\mathcal{X}_t) = \text{rank}(\mathcal{D}_t) \) and we have:
\begin{equation}
\mathcal{X}_t := U^\top \mathcal{D}_t V = \mathcal{X}_{t-1} - \eta \tilde{D}_\mathcal{B} \mathcal{X}_{t-1} \tilde{D}_\mathcal{C}.
\end{equation}
Suppose \( x_{t,ij} \) is the \( ij \) component of \( \mathcal{X}_t \), then from the equation above we have:
\begin{equation}
\tiny x_{t,ij} = x_{t-1,ij} - \eta \lambda_i \nu_j x_{t-1,ij} = (1 - \eta \lambda_i \nu_j) x_{t-1,ij} = (1 - \eta \lambda_i \nu_j)^t x_{0,ij}.
\end{equation}
Then for the first few rows \( i \) and columns \( j \) that correspond to large eigenvalues, \( x_{t,ij} \to 0 \) quickly and \( \text{rank}(\mathcal{X}_t) \) becomes small.

To make it more precise, consider the stable rank:
\begin{equation}
\text{stable-rank}(\mathcal{D}_t) = \text{stable-rank}(\mathcal{X}_t) = \frac{\| \mathcal{X}_t \|_F^2}{\| \mathcal{X}_t \|_2^2}.
\end{equation}
By the definition of the Frobenius norm, we then have:
\begin{equation}
\| \mathcal{X}_t \|_F^2 = \sum_{i=1}^m \sum_{j=1}^n (1 - \eta \lambda_i \nu_j)^{2t} x_{0,ij}^2
\end{equation}
and
\begin{equation}
\| \mathcal{X}_t \|_2^2 \geq \sum_{j=1}^n \mathcal{X}_{t,1j}^2 = \sum_{j=1}^n (1 - \eta \lambda_1 \nu_j)^{2t} x_{0,1j}^2.
\end{equation}
With high probability, \( x_{0,1j}^2 \geq \epsilon_0^2 \), since \( |x_{1i}^2| \leq c_0 \) is bounded, we have:
\begin{equation}
\text{stable-rank}(\mathcal{D}_t) \leq 1 + \frac{c_0^2}{\epsilon_0^2} \sum_{i=2}^m \frac{\sum_{j=1}^n (1 - \eta \lambda_i \nu_j)^{2t}}{\sum_{j=1}^n (1 - \eta \lambda_1 \nu_j)^{2t}}.
\end{equation}
Using Mediant inequality, \( \frac{a}{b} \leq \frac{a+c}{b+d} \leq \frac{c}{d} \) for \( a, b, c, d > 0 \), therefore, we know that for \( i \)-th row \( (i \geq 2) \), since \( \lambda_i \geq \lambda_1 \):
\begin{equation}
\tiny \frac{\sum_{j=1}^n (1 - \eta \lambda_i \nu_j)^{2t}}{\sum_{j=1}^n (1 - \eta \lambda_1 \nu_j)^{2t}} \leq \max_j \left( \frac{1 - \eta \lambda_i \nu_j}{1 - \eta \lambda_1 \nu_j} \right)^{2t} = \left( \frac{1 - \eta \lambda_i \nu_1}{1 - \eta \lambda_1 \nu_1} \right)^{2t} \leq 1
\end{equation}
and the conclusion follows. $\quad \square$

\section{Normalized Subspace}
A normalized subspace in linear algebra refers to a subspace of a vector space that has been scaled or normalized such that its vectors have a unit length. In other words, all vectors within the subspace have been adjusted such that their Euclidean norms are equal to 1. Mathematically, let $\tilde{V}$ be a vector space and $\tilde{W}$ be a subspace of $\tilde{V}$. $\tilde{W}$ is considered a normalized subspace if for every vector $ \mathbf{\tilde{v}}$ in $\tilde{W}$, the norm of $\mathbf{\tilde{v}}$ is 1, i.e., $\| \mathbf{\tilde{v}} \| = 1$. Normalized subspaces are commonly encountered in various mathematical and computational contexts, such as in optimization algorithms, signal processing, and machine learning, where ensuring consistency or stability in vector magnitudes is important. 

\begin{figure}[t!]
{\includegraphics[width=0.95\linewidth]{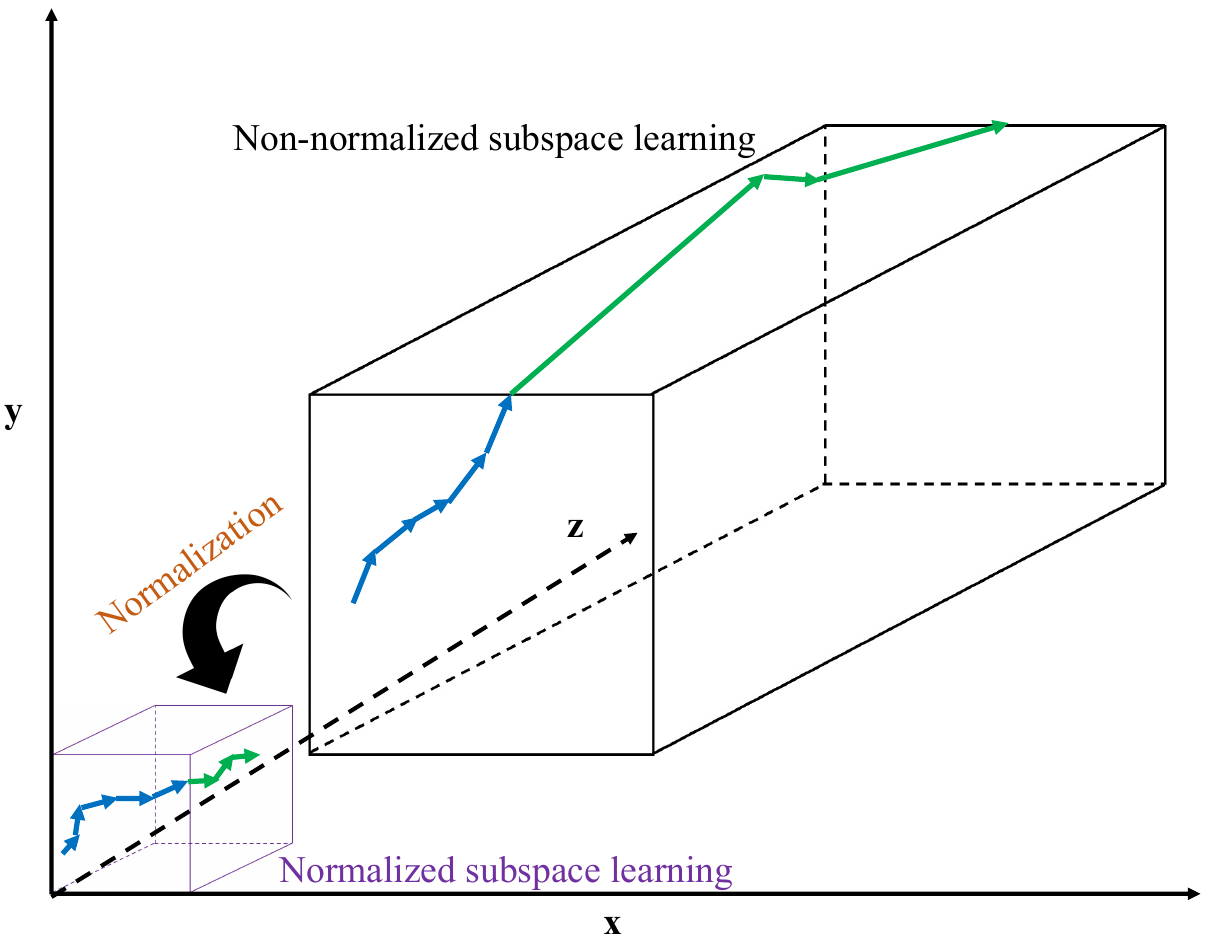}}
\caption{\footnotesize 
The diagram shows gradient descent during LLM fine-tuning in different subspaces. Fine-tuning in unnormalized subspaces (top) leads to unstable and erratic convergence, which slows down training. In contrast, normalized subspaces (bottom) result in smoother and more stable convergence, improving training efficiency. Arrow lengths represent step sizes, and different colors show learning paths in different subspaces.}
\label{fig:figure0}
\vspace{-0.2cm}
\end{figure}

\section{Asymptotic Analysis for Existing PEFT Methods with Big-O Notation}
\label{preliminary:2-1}
Asymptotic analysis is a mathematical method used to analyze the behavior of functions as their input values approach certain limits, typically either infinity or zero. This analysis is often expressed using Big-O notation $O$ to describe the worst-case time complexity or space complexity of an algorithm as a function of the size of its input. It provides an upper bound on the rate of growth of the algorithm's resource requirements as the input size increases. 

\noindent\textbf{Definition 2.1 (Big-O Notation).}
\label{big_oh}
In mathematics, the Big-O notation $O$, also known as Landau notation \cite{knuth1976big}, is used to describe the limiting behavior of a function when its argument approaches a particular value or infinity. Formally, let $f(x)$ and $g(x)$ be two functions, where $x \in \mathbb{R}$. We say that $f(x) = O(g(x))$ as $x$ approaches a certain value or infinity if and only if there exist positive constants $\tau$ and $k$ such that $|f(x)| \leq \tau \cdot |g(x)|$, $\forall x > k$, i.e., for sufficiently large $x$. This essentially means that $f(x)$ is bounded above by a constant multiple of  $g(x)$ for large enough $x$. 

\vspace{+2pt}\noindent\textbf{Theorem 2.1 (Sum of Functions).}
\label{thm1}
\textit{If $f \in O(g)$ and $f' \in O(g')$, then $f + f' \in O(g + g')$.} 

\vspace{+2pt}\noindent\textbf{Corollary 2.1 (Sum of Functions).}
\label{corollary1}
\textit{If $f \in O(g)$, $f' \in O(g')$, and $g = g'$, then $f + f' \in O(g)$.} 

\vspace{+2pt}\noindent\textbf{Theorem 2.2 (Product of Functions).}
\label{thm2}
\textit{If $f \in O(g)$ and $f' \in O(g')$, then $f \cdot f' \in O(g \cdot g')$.} 

\vspace{+2pt}\noindent\textbf{Theorem 2.3 (Composition of Functions).}
\label{thm3}
\textit{If \( f \in O(g) \) and \( f' \in O(g') \), then \( f \circ f' \in O(g \circ g') \).} 

By employing asymptotic analysis, we can compare the efficiency of the two categories of PEFT methods, which primarily differ in whether they modify a given model's architecture. Based on their practical implementations, we can conceptualize the category involving changes to the model architecture as the composition of functions, while the other category can be viewed as the sum of functions. According to the above two theorems, the complexity of the composition of functions is much larger than the sum of functions in the worst-case scenario. In terms of efficiency during the inference phase, this discrepancy highlights the general preference for PEFT methods that avoid modifying a model's architecture. Proofs of the theorems are provided as follows.

\vspace{+2pt}\noindent\textbf{Theorem 2.1 (Sum of Functions).}
\label{appendix:thms}
\textit{If $f \in O(g)$ and $f' \in O(g')$, then $f + f' \in O(g + g')$.}

\noindent\textbf{\textit{<Proof>}}
To prove the statement, we need to use the \textit{``Definition 2.1 Big-O Notation''} (\ref{big_oh}) and show that there exist constants \( C \) and \( n_0 \) such that \( |f(n) + f'(n)| \leq C|g(n) + g'(n)| \) for all \( n \geq n_0 \).

Given \( f(n) \in O(g(n)) \) and \( f'(n) \in O(g'(n)) \), by \textit{``Definition 2.1 Big-O Notation''} (\ref{big_oh}), the above statement means:

i. There exist constants \( C_1 > 0 \) and \( n_1 \) such that for all \( n \geq n_1 \):
\begin{equation}
    |f(n)| \leq C_1 |g(n)|.
\end{equation}

ii. There exist constants \( C_2 > 0 \) and \( n_2 \) such that for all \( n \geq n_2 \):
\begin{equation}
    |f'(n)| \leq C_2 |g'(n)|.
\end{equation}
   
We need to show that \( f(n) + f'(n) \in O(g(n) + g'(n)) \). Specifically, we want to find constants \( C > 0 \) and \( n_0 \) such that for all \( n \geq n_0 \):
\begin{equation}
    |f(n) + f'(n)| \leq C |g(n) + g'(n)|.
\end{equation}

Consider the absolute value of the sum \( f(n) + f'(n) \):
\begin{equation}
    |f(n) + f'(n)| \leq |f(n)| + |f'(n)|.
\end{equation}

Using the given conditions \( |f(n)| \leq C_1 |g(n)| \) and \( |f'(n)| \leq C_2 |g'(n)| \), we have:
\begin{equation}
    |f(n) + f'(n)| \leq |f(n)| + |f'(n)| \leq C_1 |g(n)| + C_2 |g'(n)|.
\end{equation}

Now, observe that:
\begin{equation}
    \tiny C_1 |g(n)| + C_2 |g'(n)| \leq (C_1 + C_2) \max(|g(n)|, |g'(n)|) \leq (C_1 + C_2) |g(n) + g'(n)|.
\end{equation}

Therefore, we can set \( C = C_1 + C_2 \). Let \( n_0 = \max(n_1, n_2) \), which ensures that both conditions \( |f(n)| \leq C_1 |g(n)| \) and \( |f'(n)| \leq C_2 |g'(n)| \) hold for all \( n \geq n_0 \).

Hence, for all \( n \geq n_0 \):
\begin{equation}
    |f(n) + f'(n)| \leq (C_1 + C_2) |g(n) + g'(n)|.
\end{equation}

This demonstrates that \( f(n) + f'(n) \in O(g(n) + g'(n)) \).

Therefore, we have proved that if \( f \in O(g) \) and \( f' \in O(g') \), then \( f + f' \in O(g + g') \). $\quad \square$

\vspace{+2pt}\noindent\textbf{Corollary 2.1 (Sum of Functions).}
\textit{If $f \in O(g)$, $f' \in O(g')$, and $g = g'$, then $f + f' \in O(g)$.}

\noindent\textbf{\textit{<Proof>}}
To prove the statement, we need to use the \textit{``Definition 2.1 Big-O Notation''} (\ref{big_oh}) and show that there exist constants \( C \) and \( n_0 \) such that \( |f(n) + f'(n)| \leq C|g(n)| \) for all \( n \geq n_0 \).

Given \( f(n) \in O(g(n)) \) and \( f'(n) \in O(g(n)) \) (since \( g = g' \)), by \textit{``Definition 2.1 Big-O Notation''} (\ref{big_oh}), the above statement means:

i. There exist constants \( C_1 > 0 \) and \( n_1 \) such that for all \( n \geq n_1 \):
\begin{equation}
    |f(n)| \leq C_1 |g(n)|.
\end{equation}

ii. There exist constants \( C_2 > 0 \) and \( n_2 \) such that for all \( n \geq n_2 \):
\begin{equation}
    |f'(n)| \leq C_2 |g(n)|.
\end{equation}

We need to show that \( f(n) + f'(n) \in O(g(n)) \). Specifically, we want to find constants \( C > 0 \) and \( n_0 \) such that for all \( n \geq n_0 \):
\begin{equation}
    |f(n) + f'(n)| \leq C |g(n)|.
\end{equation}

Consider the absolute value of the sum \( f(n) + f'(n) \):
\begin{equation}
    |f(n) + f'(n)| \leq |f(n)| + |f'(n)|.
\end{equation}

Using the given conditions \( |f(n)| \leq C_1 |g(n)| \) and \( |f'(n)| \leq C_2 |g(n)| \), we have:
\begin{equation}
    |f(n) + f'(n)| \leq |f(n)| + |f'(n)| \leq C_1 |g(n)| + C_2 |g(n)|.
\end{equation}

Now, observe that:
\begin{equation}
    C_1 |g(n)| + C_2 |g(n)| = (C_1 + C_2) |g(n)|.
\end{equation}

Therefore, we can set \( C = C_1 + C_2 \). Let \( n_0 = \max(n_1, n_2) \), which ensures that both conditions \( |f(n)| \leq C_1 |g(n)| \) and \( |f'(n)| \leq C_2 |g(n)| \) hold for all \( n \geq n_0 \).

Hence, for all \( n \geq n_0 \):
\begin{equation}
    |f(n) + f'(n)| \leq (C_1 + C_2) |g(n)|.
\end{equation}

This demonstrates that \( f(n) + f'(n) \in O(g(n)) \).

Therefore, we have proved that if \( f \in O(g) \), \( f' \in O(g') \), and \( g = g' \), then \( f + f' \in O(g) \). $\quad \square$

\vspace{+2pt}\noindent\textbf{Theorem 2.2 (Product of Functions).}
\textit{If $f \in O(g)$ and $f' \in O(g')$, then $f \cdot f' \in O(g \cdot g')$.}

\noindent\textbf{\textit{<Proof>}} To prove the statement, we need to use the \textit{``Definition 2.1 Big-O Notation''} (\ref{big_oh}) and show that there exist constants \( C \) and \( n_0 \) such that \( |f(n) \cdot f'(n)| \leq C|g(n) \cdot g'(n)| \) for all \( n \geq n_0 \).

Given

i. \( f(n) \in O(g(n)) \): There exist constants \( C_1 > 0 \) and \( n_1 \) such that for all \( n \geq n_1 \):
\begin{equation}
    |f(n)| \leq C_1 |g(n)|.
\end{equation}

ii. \( f'(n) \in O(g'(n)) \): There exist constants \( C_2 > 0 \) and \( n_2 \) such that for all \( n \geq n_2 \):
\begin{equation}
    |f'(n)| \leq C_2 |g'(n)|.
\end{equation}

We need to show that \( f(n) \cdot f'(n) \in O(g(n) \cdot g'(n)) \). Specifically, we want to find constants \( C > 0 \) and \( n_0 \) such that for all \( n \geq n_0 \):
\begin{equation}
    |f(n) \cdot f'(n)| \leq C |g(n) \cdot g'(n)|.
\end{equation}

Using the given conditions, we have for all \( n \geq n_0 = \max(n_1, n_2) \):
\begin{equation}
    |f(n)| \leq C_1 |g(n)|,
\end{equation}
\begin{equation}
    |f'(n)| \leq C_2 |g'(n)|.
\end{equation}

Now, consider the product \( |f(n) \cdot f'(n)| \):
\begin{equation}
    |f(n) \cdot f'(n)| = |f(n)| \cdot |f'(n)|.
\end{equation}

Using the inequalities from the given conditions, we get:
\begin{equation}
    |f(n) \cdot f'(n)| \leq (C_1 |g(n)|) \cdot (C_2 |g'(n)|).
\end{equation}

This simplifies to:
\begin{equation}
    |f(n) \cdot f'(n)| \leq C_1 C_2 |g(n)| \cdot |g'(n)|.
\end{equation}

Let \( C = C_1 C_2 \). Thus, we have:
\begin{equation}
    |f(n) \cdot f'(n)| \leq C |g(n) \cdot g'(n)|.
\end{equation}

Therefore, we have shown that if \( f \in O(g) \) and \( f' \in O(g') \), then \( f \cdot f' \in O(g \cdot g') \). $\quad \square$

\vspace{+2pt}\noindent\textbf{Theorem 2.3 (Composition of Functions).}
\textit{If \( f \in O(g) \) and \( f' \in O(g') \), then \( f \circ f' \in O(g \circ g') \).}

\noindent\textbf{\textit{<Proof>}} To prove the statement, we need to show that there exist constants \( C \) and \( n_0 \) such that \( |(f \circ f')(n)| \leq C|(g \circ g')(n)| \) for all \( n \geq n_0 \).

Given

i. \( f(n) \in O(g(n)) \): There exist constants \( C_1 > 0 \) and \( n_1 \) such that for all \( n \geq n_1 \):
\begin{equation}
    |f(n)| \leq C_1 |g(n)|.
\end{equation}

ii. \( f'(n) \in O(g'(n)) \): There exist constants \( C_2 > 0 \) and \( n_2 \) such that for all \( n \geq n_2 \):
\begin{equation}
    |f'(n)| \leq C_2 |g'(n)|.
\end{equation}

We need to show that \( f(f'(n)) \in O(g(g'(n))) \). Specifically, we want to find constants \( C > 0 \) and \( n_0 \) such that for all \( n \geq n_0 \):
\begin{equation}
    |f(f'(n))| \leq C |g(g'(n))|.
\end{equation}

Using the given conditions, we know that for \( f'(n) \) when \( n \geq n_2 \):
\begin{equation}
    |f'(n)| \leq C_2 |g'(n)|.
\end{equation}

This inequality tells us that \( f'(n) \) is bounded by a multiple of \( g'(n) \).

Now, consider \( f(f'(n)) \). By the \textit{``Definition 2.1 Big-O Notation''} (\ref{big_oh}), for \( f(x) \) when \( x \geq n_1 \):
\begin{equation}
    |f(x)| \leq C_1 |g(x)|.
\end{equation}

We need to substitute \( f'(n) \) for \( x \):
\begin{equation}
    |f(f'(n))| \leq C_1 |g(f'(n))|.
\end{equation}

We now apply the bound for \( f'(n) \):
\begin{equation}
    |f(f'(n))| \leq C_1 |g(C_2 g'(n))|.
\end{equation}

Since \( g \in O(g) \), scaling the argument by a constant factor does not change the Big-O classification. Therefore:
\begin{equation}
    |g(C_2 g'(n))| \leq C_3 |g(g'(n))|.
\end{equation}
for some constant \( C_3 > 0 \). This follows from the property of functions within Big-O notation where scaling the input by a constant does not affect the overall classification.

Hence, we have:
\begin{equation}
    |f(f'(n))| \leq C_1 C_3 |g(g'(n))|.
\end{equation}

Let \( C = C_1 C_3 \). Thus:
\begin{equation}
    |f(f'(n))| \leq C |g(g'(n))|.
\end{equation}

Therefore, we have shown that if \( f \in O(g) \) and \( f' \in O(g') \), then \( f \circ f' \in O(g \circ g') \). $\quad \square$

\section{Proof of Theorem 4 (\cite{zhao2024galore})}
\noindent\textbf{Theorem 4 (Gradient Form of Reversible Models).}
\textit{In a chained reversible neural network \( N(\mathbf{x}) := N_L(N_{L-1}(\ldots N_1(\mathbf{x}))) \) with \(\ell_2\)-objective \(\phi := \frac{1}{2} \lVert \mathbf{y} - N(\mathbf{x}) \rVert_2^2\), the weight matrix \( \mathcal{W}_l \) at layer \( l \) has gradient \( \mathcal{D}_l \) of the following form for batchsize 1:
\begin{equation}
    \mathcal{D}_l = [J_l^\top \mathbf{y} \mathbf{f}_{l-1}^\top] - [J_l^\top J_l] \mathcal{W}_l [\mathbf{f}_{l-1} \mathbf{f}_{l-1}^\top],
\end{equation}
where \( J_l := \textup{Jacobian}(N_L) \ldots \textup{Jacobian}(N_{l+1}) \) and \( \mathbf{f}_l := N_l(N_{l-1} \ldots N_1(\mathbf{x})) \).}

\noindent
\textbf{\textit{<Proof>}} Note that for a layered reversible network, we have
\begin{equation}
    N(\mathbf{x}) = N_L(N_{L-1}(\ldots N_1(\mathbf{x}))) = \mathcal{K}_L(\mathbf{x})\mathcal{K}_{L-1}(\mathbf{x})\ldots \mathcal{K}_1(\mathbf{x})\mathbf{x}.
\end{equation}
Let $\mathbf{f}_l := N_l(N_{l-1}(\ldots N_1(\mathbf{x})))$ and $J_l := \mathcal{K}_L(\mathbf{x})\ldots \mathcal{K}_{l+1}(\mathbf{x})$, and for linear layer $l$, we can write $N(\mathbf{x}) = J_l \mathcal{W}_l \mathbf{f}_{l-1}$.
Therefore, for the linear layer $l$ with weight matrix $\mathcal{W}_l$, we have:
\begin{gather}
    \scriptsize d\phi = (\mathbf{y}  - N(\mathbf{x}))^\top dN(\mathbf{x}), \nonumber \\
    \scriptsize = (\mathbf{y}  - N(\mathbf{x}))^\top K_L(\mathbf{x})\ldots K_{l+1} (\mathbf{x}) d\mathcal{W}_l \mathbf{f}_{l-1} + \text{terms not related to } d\mathcal{W}_l, \nonumber \\
    = (\mathbf{y}  - J_l \mathcal{W}_l \mathbf{f}_{l-1})^\top J_l d\mathcal{W}_l \mathbf{f}_{l-1}, \nonumber \\ 
    = \text{tr}(d\mathcal{W}_l^\top J_l^\top (\mathbf{y}  - J_l \mathcal{W}_l \mathbf{f}_{l-1}) \mathbf{f}_{l-1}^\top). 
\end{gather}
This gives the gradient of $\mathcal{W}_l$:
\begin{equation}
    \mathcal{D}_l = J_l^\top \mathbf{y}  \mathbf{f}_{l-1}^\top - J_l^\top J_l \mathcal{W}_l \mathbf{f}_{l-1} \mathbf{f}_{l-1}^\top. 
\end{equation}
$\quad \square$

For the softmax objective with small logits, we can similarly show that the backpropagated gradient maintains the same structure. Consequently, \textbf{Theorem 4} is also applicable in this context \cite{zhao2024galore}.

\section{Proof of Lemma 2 (\cite{zhao2024galore})}
\noindent\textbf{Lemma 2 (Gradient Structure of Softmax Loss).} 
\textit{For \( \kappa \)-way logsoftmax loss \(\phi(\mathbf{y}; \mathbf{f}) := - \log \left( \frac{\exp(\mathbf{y}^\top \mathbf{f})}{\mathbf{1}^\top \exp(\mathbf{f})} \right)\), let \( \hat{\mathbf{f}} = \tilde{P}_\mathbf{1}^\perp \mathbf{f} \) be the zero-mean version of network output \( \mathbf{f} \), where \( \tilde{P}_\mathbf{1}^\perp := I - \frac{1}{\kappa} \mathbf{1} \mathbf{1}^\top \), then we have:
\begin{equation}
    -d\phi = \mathbf{y}^\top d\hat{\mathbf{f}} - \frac{\gamma \hat{\mathbf{f}}^\top d\hat{\mathbf{f}}}{\kappa} + O\left( \frac{\hat{\mathbf{f}}^2}{\kappa} \right) d\hat{\mathbf{f}},
\end{equation}
where \( \gamma(\mathbf{y}, \mathbf{f}) \approx 1 \) and \( \mathbf{y} \) is a data label with \( \mathbf{y}^\top \mathbf{1} = 1 \).}

\textbf{\textit{<Proof>}} 
Let $\hat{\mathbf{f}} := \tilde{P}_\mathbf{1}^\perp \mathbf{f}$ be the zero-mean version of network output $\mathbf{f}$. Then we have $\mathbf{1}^\top \hat{\mathbf{f}} = 0$ and $\mathbf{f} = \hat{\mathbf{f}} + c\mathbf{1}$. Therefore, we have:
\begin{equation}
    -\phi = \log \left( \frac{\exp(c) \exp(\mathbf{y}^\top \hat{\mathbf{f}})}{\exp(c) \mathbf{1}^\top \exp(\hat{\mathbf{f}})} \right) 
= \mathbf{y}^\top \hat{\mathbf{f}} - \log(\mathbf{1}^\top \exp(\hat{\mathbf{f}})).
\end{equation}
Using the Taylor expansion $\exp(x) = 1 + x + \frac{x^2}{2} + o(x^2)$, we have:
\begin{equation}
    \mathbf{1}^\top \exp(\hat{\mathbf{f}}) = \mathbf{1}^\top (\mathbf{1} + \hat{\mathbf{f}} + \frac{1}{2} \hat{\mathbf{f}}^2) + o(\hat{\mathbf{f}}^2) = K \left(1 + \frac{\hat{\mathbf{f}}^\top \hat{\mathbf{f}}}{2\kappa} + o\left(\frac{\hat{\mathbf{f}}^2}{\kappa}\right) \right).
\end{equation}
So,
\begin{equation}
    -\phi = \mathbf{y}^\top \hat{\mathbf{f}} - \log\left(1 + \frac{\hat{\mathbf{f}}^\top \hat{\mathbf{f}}}{2\kappa} + o\left(\frac{\hat{\mathbf{f}}^2}{\kappa}\right)\right) - \log \kappa.
\end{equation}
Therefore,
\begin{equation}
    -d\phi = \mathbf{y}^\top d\hat{\mathbf{f}} - \frac{\tilde{\gamma}}{\kappa} \hat{\mathbf{f}}^\top d\hat{\mathbf{f}} + O\left(\frac{\hat{\mathbf{f}}^2}{\kappa}\right) d\hat{\mathbf{f}},
\end{equation}
where $\tilde{\gamma} := \left(1 + \frac{\hat{\mathbf{f}}^\top \hat{\mathbf{f}}}{2\kappa} + o\left(\frac{\hat{\mathbf{f}}^2}{\kappa}\right)\right)^{-1} \approx 1$.
$\quad \square$

With this lemma, it is clear that for a reversible network \( \mathbf{f} := N(\mathbf{x}) = J_l(\mathbf{x})\mathcal{W}_l\mathbf{f}_{l-1}(\mathbf{x}) \), the gradient \( \mathcal{D}_l \) of \( \mathcal{W}_l \) has the following form:
\begin{equation}
    \mathcal{D}_l = [J_l \tilde{P}_\mathbf{1}^\perp \mathbf{y} \mathbf{f}_{l-1}] - [\tilde{\gamma} J_l^\top \tilde{P}_\mathbf{1}^\perp J_l] \mathcal{W}_l [\frac{\mathbf{f}_{l-1} \mathbf{f}_{l-1}^\top}{\kappa}],
\end{equation}
which is consistent with the form \( \mathcal{D}_l = \mathcal{A} - \mathcal{B}\mathcal{W}_l\mathcal{C} \). Refer to \cite{tian2020understanding} for a comprehensive introduction to reversibility.


\begin{figure*}[t!]
    \centering
    \scriptsize
    \begin{subfigure}{0.3\textwidth}
        \centering
        \begin{adjustbox}{width=\textwidth} 
            \begin{tikzpicture}
                \begin{axis}[
                    width=1.3\textwidth, 
                    height=1.3\textwidth, 
                    bar width=3pt, 
                    ymin=0,
                    ymax=50,
                    ybar,
                    symbolic x coords={350M, 1B, 3B, 7B},
                    xtick=data,
                    ylabel={Memory cost (GB)},
                    xlabel={Model Size},
                    legend style={at={(0.5,1.07)}, anchor=south, legend columns=2},
                    enlarge x limits=0.15,
                    enlarge y limits=0.1
                ]
                \addplot[fill=gray] coordinates {(350M,3.75) (1B,13.17) (3B,27.79) (7B,0)};
                \addplot[fill=brown] coordinates {(350M,2.67) (1B,9.96) (3B,18.91) (7B,47.23)};
                \addplot[fill=orange, postaction={pattern=north west lines}] coordinates {(350M,1.95) (1B,6.59) (3B,10.02) (7B,22.19)};
                \addplot[fill=black, postaction={pattern=grid}] coordinates {(350M,1.76) (1B,6.01) (3B,9.21) (7B,20.07)};

                \legend{BF16, 8-bit Adam, 8-bit GaLore, 8-bit GradNormLoRP}

                \draw[red, thick, dashed] (axis cs:350M,24) -- (axis cs:7B,24) node[anchor=south east, pos=0.6, red] {RTX 4090};
                \end{axis}
            \end{tikzpicture}
        \end{adjustbox}
        \label{fig:rm}
    \end{subfigure}
    \hfill
    \begin{subfigure}{0.3\textwidth}
        \centering
        \begin{tikzpicture}
            \begin{axis}[
                height=\textwidth,
                xlabel={Update Frequency},
                ylabel={Perplexity},
                xmode=log,
                xtick={10, 100, 250, 500, 1000, 5000, 10000, 20000},
                xticklabels={10, 100, 250, 500, 1000, 5000, 10000, 20000},
                xticklabel style={rotate=45, anchor=east},
                legend style={
                    at={(0.5,1.07)},
                    anchor=south,
                    column sep=1ex,
                    font=\scriptsize
                },
                legend cell align={left},
                legend columns=2
            ]
            \addplot[mark=o, color=green] coordinates {
                (10, 38.43) (100, 37.07) (250, 34.42) (500, 34.51) (1000, 35.74) (5000, 36.56) (10000, 37.51) (20000, 39.29)
            };
            \addlegendentry{Rank = 64}
            
            \addplot[mark=square, color=red] coordinates {
                (10, 35.73) (100, 34.26) (250, 31.02) (500, 31.09) (1000, 31.17) (5000, 32.42) (10000, 33.65) (20000, 35.98)
            };
            \addlegendentry{Rank = 128}
            
            \addplot[mark=triangle, color=blue] coordinates {
                (10, 33.61) (100, 28.31) (250, 26.03) (500, 25.71) (1000, 26.91) (5000, 28.35) (10000, 29.33) (20000, 30.17)
            };
            \addlegendentry{Rank = 256}
            
            \addplot[mark=diamond, color=orange] coordinates {
                (10, 32.91) (100, 25.21) (250, 25.07) (500, 25.99) (1000, 26.37) (5000, 27.33) (10000, 27.39) (20000, 28.96)
            };
            \addlegendentry{Rank = 512}
            \end{axis}
        \end{tikzpicture}
        \label{fig:freq}
    \end{subfigure}
    \hfill
    \begin{subfigure}{0.3\textwidth}
        \centering
        \begin{tikzpicture}
            \begin{axis}[
                height=\textwidth,
                xlabel={Rank},
                ylabel={Perplexity},
                xmode=log,
                xtick={32, 64, 128, 256, 512},
                xticklabels={32, 64, 128, 256, 512},
                xticklabel style={rotate=45, anchor=east},
                legend style={
                    at={(0.5, 1.08)},
                    anchor=south,
                    column sep=1ex,
                    font=\scriptsize
                },
                legend cell align={left},
                legend columns=2
            ]
            \addplot[mark=o, color=green] coordinates {
                (32, 49.37) (64, 39.76) (128, 35.76) (256, 32.62) (512, 27.31)
            };
            \addlegendentry{10k Steps}
            
            \addplot[mark=square, color=orange] coordinates {
                (32, 40.51) (64, 33.72) (128, 31.67) (256, 27.13) (512, 25.26)
            };
            \addlegendentry{20k Steps}
            
            \addplot[mark=triangle, color=blue] coordinates {
                (32, 35.25) (64, 34.19) (128, 27.61) (256, 26.19) (512, 24.23)
            };
            \addlegendentry{40k Steps}
            
            \addplot[mark=diamond, color=red] coordinates {
                (32, 34.76) (64, 33.02) (128, 27.09) (256, 25.37) (512, 23.32)
            };
            \addlegendentry{60k Steps}
            
            \addplot[mark=star, color=cyan] coordinates {
                (32, 33.89) (64, 29.12) (128, 26.33) (256, 24.61) (512, 21.54)
            };
            \addlegendentry{80k Steps}
            \end{axis}
        \end{tikzpicture}
        \label{fig:rank}
    \end{subfigure}
    \vspace{-0.6cm}
    \caption{\footnotesize From left to right, the figure illustrates a comparison of memory usage, the impact of varying subspace frequencies, and the effect of rank across steps.}
    \label{fig:combined-figures1}

\end{figure*}
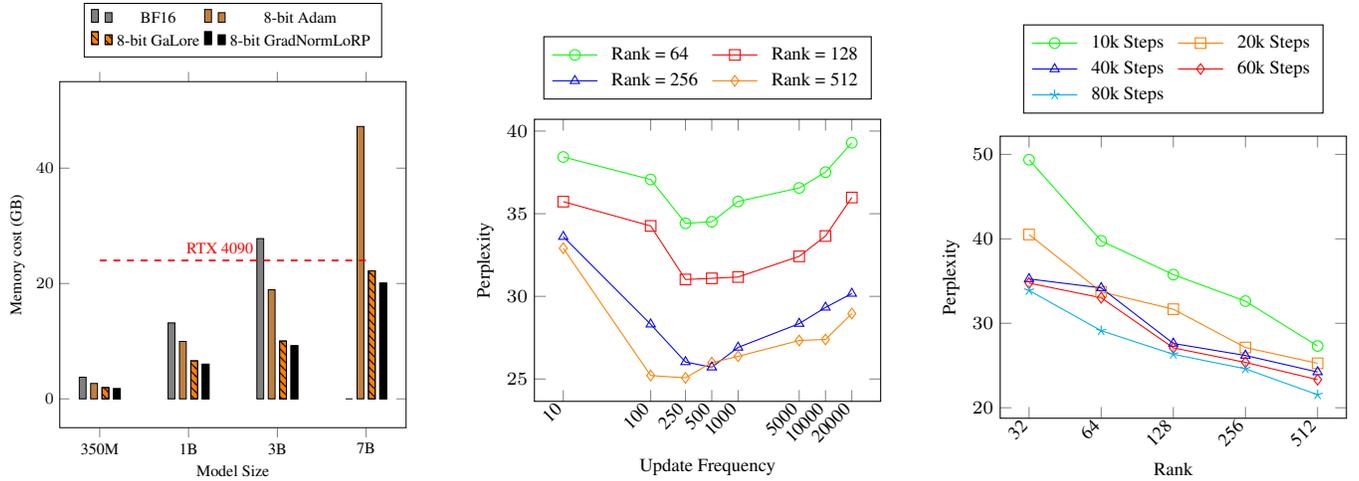

\begin{table*}[t!]
\small
\centering
\scalebox{0.75}{
\begin{tabular}{c|cc|ccccccccccc}
\toprule
\multicolumn{2}{c|}{\textbf{Model}}  &\textbf{Memory} & \textbf{CoLA} & \textbf{MNLI} & \textbf{MRPC} & \textbf{QNLI} & \textbf{QQP} & \textbf{RTE} & \textbf{SST-2} & \textbf{STS-B} & \textbf{WNLI} & \textbf{Avg}\\ 
\midrule
\multicolumn{2}{c|}{Full FT$_\textup{base}$} &747M  &57.03 & 86.30 & 92.09 & 92.04 & 91.35 & 77.62 & 91.74 & 90.82 & 43.90  & 80.32  \\ 
\multicolumn{2}{c|}{Full FT$_\textup{large}$}  & 1.66G & 58.39 & 89.05 & 92.86 & 94.42 & 91.37 & 83.36 & 95.21 & 92.31 & 44.55  & 82.39  \\ 
\midrule
\multicolumn{2}{c|}{Ours$_\textup{base}$(r=4)}  & 249M & 59.31 & 86.42	& 91.10	& 92.48	& 90.60 & 75.81	& 94.50 & 90.85 & \textbf{47.89} & 81.01\\ 
\rowcolor{mygray} \multicolumn{2}{c|}{Ours$_\textup{large}$ (r=4)}  & 495M & \textbf{63.31} & \textbf{89.02} & \textbf{92.25} & \textbf{94.14} & \textbf{90.91} & \textbf{85.20} & \textbf{94.72} & \textbf{91.95} & 35.21  & \textbf{81.86}  \\ 
\midrule
\multicolumn{2}{c|}{Ours$_\textup{base}$(r=8)}  & 251M & 60.31 & 86.97	& 91.36	& 92.62	& 91.01 & 77.26	& 94.50 & 90.83  & \textbf{40.85} & 80.63\\ 
\rowcolor{mygray} \multicolumn{2}{c|}{Ours$_\textup{large}$(r=8)}  & 534M & \textbf{62.08} & \textbf{89.13} & \textbf{92.86} & \textbf{94.36} & \textbf{92.07} & \textbf{83.39} & \textbf{95.53} & \textbf{91.62} & \textbf{40.85}  & \textbf{82.43}  \\ 
\bottomrule
\end{tabular}}
\caption{\footnotesize Ablation study for model size under the same model architecture. ``r'' denotes rank, ``Ours'' indicates GradNormLoRP, and ``FT'' signifies fine-tuning.}
\label{table:2-1}
\vspace{-0.2cm}
\end{table*}

\begin{table*}[t!]
\small
\centering
\scalebox{0.75}{
\begin{tabular}{c|cc|ccccccccccc}
\toprule
\multicolumn{2}{c|}{\textbf{Model}}  &\textbf{Memory} & \textbf{CoLA} & \textbf{MNLI} & \textbf{MRPC} & \textbf{QNLI} & \textbf{QQP} & \textbf{RTE} & \textbf{SST-2} & \textbf{STS-B} & \textbf{WNLI} & \textbf{Avg}\\ 
\midrule
\multicolumn{2}{c|}{Full FT}  & 747M & 57.03 & 86.30 & 92.09 & 92.04 & 91.35 & 77.62 & 91.74 & 90.82 &43.90  & 80.32  \\ 
\midrule
\multicolumn{2}{c|}{w/o norm (r=4)}   & 248M & 58.80	& \textbf{86.43} & 90.43	& 92.49	& 90.41	& \textbf{76.17} & \textbf{94.61}	& 90.67 & 46.48  & 80.72 \\ 
\rowcolor{mygray} \multicolumn{2}{c|}{w/ norm (r=4)}  & 249M & \textbf{59.31} & 86.42	& \textbf{91.10}	& \textbf{92.48}	& \textbf{90.60} & 75.81	& 94.50 & \textbf{90.85} & \textbf{47.89} & \textbf{81.01}\\ 
\midrule
\multicolumn{2}{c|}{w/o norm (r=8)}   & 250M & \textbf{61.57}	& 86.79 & \textbf{92.14}	& 92.51	& 90.90	& 76.90 & 94.27	& 90.64 & 36.62  & 80.26 \\  
\rowcolor{mygray} \multicolumn{2}{c|}{w/ norm (r=8)}  & 251M & 60.31 & \textbf{86.97}	& 91.36	& \textbf{92.62}	& \textbf{91.01} & \textbf{77.26}	& \textbf{94.50} & \textbf{90.83}  & \textbf{40.85} & \textbf{80.65}\\ 
\bottomrule
\end{tabular}}
\caption{\footnotesize Ablation study for weight normalization. ``r'' denotes rank, ``norm'' indicates normalization, and ``FT'' signifies fine-tuning.}
\label{table:2-1wo}
\vspace{-0.2cm}
\end{table*}

\begin{table*}[t!]
\small
\centering
\scalebox{0.75}{
\begin{tabular}{c|cc|ccccccccccc}
\toprule
\multicolumn{2}{c|}{\textbf{Model}}  &\textbf{Memory} & \textbf{CoLA} & \textbf{MNLI} & \textbf{MRPC} & \textbf{QNLI} & \textbf{QQP} & \textbf{RTE} & \textbf{SST-2} & \textbf{STS-B} & \textbf{WNLI} & \textbf{Avg}\\ 
\midrule
\multicolumn{2}{c|}{Full FT}  & 747M & 57.03 & 86.30 & 92.09 & 92.04 & 91.35 & 77.62 & 91.74 & 90.82 & 43.90  & 80.32  \\ 
\midrule
\multicolumn{2}{c|}{w/o GP (r=4)}      & 259M & 55.51	& \textbf{86.90} & 89.91	& 92.24	& 89.54& 70.75 & 93.46	& 90.04 & 56.34  & 80.52 \\  
\rowcolor{mygray} \multicolumn{2}{c|}{w/ GP (r=4)}  & 249M & \textbf{59.31} & 86.42	& \textbf{91.10}	& \textbf{92.48}	& \textbf{90.60} & 75.81	& 94.50 & \textbf{90.85} & \textbf{47.89} & \textbf{81.01}\\ 
\midrule
\multicolumn{2}{c|}{w/o GP (r=8)}      & 267M & 57.27	& 86.55 & 89.56	& 92.46	& 89.86	& 70.04 & \textbf{94.15}	& 90.16 & \textbf{43.66}  & 79.30 \\ 
\rowcolor{mygray} \multicolumn{2}{c|}{w/ GP (r=8)}  & 251M &\textbf{60.31} & \textbf{86.97}	& 91.36	& \textbf{92.62}	& \textbf{91.01} & 77.26	& \textbf{94.50} & \textbf{90.83}  & 40.85 & \textbf{80.65}\\

\bottomrule
\end{tabular}}
\caption{\footnotesize Investigating the impact of gradient projection through ablation. ``r'' denotes rank, ``GP'' indicates gradient projection, and ``FT'' signifies fine-tuning.}
\label{table:2-1wop}
\vspace{-0.2cm}
\end{table*}

\end{document}